\documentclass{article}

\usepackage{arxiv}
\usepackage{subcaption}
\usepackage{comment}
\usepackage{graphicx} 
\usepackage{svg}

\usepackage{float} 
\usepackage[utf8]{inputenc} 
\usepackage[T1]{fontenc}    
\usepackage{hyperref}       
\usepackage{url}            
\usepackage{booktabs}       
\usepackage{amsfonts}       
\usepackage{amsmath}
\usepackage[sort]{cite}
\usepackage{nicefrac}       
\usepackage{microtype}      
\usepackage{xcolor}         
\usepackage{graphicx} 
\usepackage{svg}
\usepackage{subcaption}
\usepackage{todonotes}
\usepackage{comment}
\usepackage{placeins}

\title{The Mean Dimension of Neural Networks - What causes the interaction effects?}

\author{Roman Hahn \\
	Department of Decision Sciences\\
	Bocconi University\\
	Via Roentgen 1, 20136 Milan, Italy\ \\
	\texttt{roman.hahn@unibocconi.it} \\
   \And
  Christoph Feinauer\\
  Department of Computing Sciences \\ 
  Bocconi University \\ 
  Via Roentgen 1, 20136 Milan, Italy\\
  \texttt{christoph.feinauer@unibocconi.it}
   \And
  Emanuele Borgonovo\thanks{corresponding author}\\
  Department of Decision Sciences\\
  Bocconi University\\
  Via Roentgen 1, 20136 Milan, Italy\\
  \texttt{emanuele.borgonovo@unibocconi.it}
}

\begin{document}
\maketitle

\begin{abstract}
Owen and Hoyt recently showed  that the effective dimension offers key structural information about the input-output mapping underlying an artificial neural network. Along this line of research, this work proposes an estimation procedure that allows the calculation of the mean dimension from a given dataset, without resampling from external distributions. The design yields total indices when features are independent and a variant of total indices when features are correlated. We show that this variant possesses the zero independence property. With synthetic datasets, we analyse how the mean dimension evolves layer by layer and how the activation function impacts the magnitude of interactions. We then use the mean dimension to study some of the most widely employed convolutional architectures for image recognition (LeNet, ResNet, DenseNet). To account for pixel correlations, we propose calculating the mean dimension after the addition of an inverse PCA layer that allows one to work on uncorrelated PCA-transformed features, without the need to retrain the neural network. We use the generalized total indices to produce heatmaps for post-hoc explanations, and we employ the mean dimension on the PCA-transformed features for cross comparisons of the artificial neural networks structures. Results provide several insights on the difference in magnitude of interactions across the architectures, as well as indications on how the mean dimension evolves during training.
\end{abstract}

\section{Introduction}

While the success of machine learning models based on deep artificial neural networks has revolutionized fields like computer vision \cite{voulodimos2018deep},
natural language processing \cite{young2018recent} and reinforcement learning \cite{li2017deep}, there is currently no theory that can explain their empirically observed generalization performance or give a satisfactory answer to the question on how it is achieved \cite{Sejnowski2020}. In addition, the large computational resources needed for training and inference of state-of-the-art networks often prohibit ablation studies, hinder comparison of models and render the application of standard machine learning tools like cross-validation difficult.

Explainability and interpretability of black-box artificial neural networks are topical research subjects. The review of \cite{GuidMonr18} analyzes various needs for interpretability and explainability in complex machine learning tasks. The work \cite{Begoli2019} emphasizes the need of uncertainty quantification in deep learning. 
Rudin \cite{Rudin2019} recommends to favour interpretable models over black-boxes when the problem permits. The review of \cite{RudiChen22}  clarifies further the distinction between interpretability and explainability: the former refers to the use machine learning models which are as transparent as possible, the latter to the derivation  of post-hoc insights that analysts can use to corroborate results and to make inference more robust.

Methods for explainability can be broadly divided into two categories: The first comprises tools that act at the individual prediction level. Techniques such as LIME \cite{Ribeiro2016} or Layerwise Relevance Propagation \cite{Bach2015} belong to this category. The second category comprises dataset level methods. To this group belong Feature Permutation \cite{Breiman2001} or Shapley-values \cite{Lundberg2017} or measures of statistical association \cite{GambKlei18}. 


Artificial neural networks are often regarded as black-boxes: While the single neuron is a mathematically simple unit, the input-output relationship that results from the
combination of millions of such units is a highly complex function. This, in turn, makes the \textit{interpretation} and \textit{explanation} of artificial neural network decisions hard \cite{Montavon2018}, 
making stakeholders reluctant in accepting advice from models that are fundamentally opaque \cite{london2019artificial}.

A solid background to obtain insights on black-box models is represented by the theory of the functional ANOVA expansion \cite{EfroStei81,Owen03,BarrRabi22}. This expansion allows analysts to decompose the prediction variance as a sum of variance contributions by all subgroups of indices. 
In \cite{CaflMoro97} and \cite{Owen03}, Owen introduces the notion of mean dimension. Intuitively, the mean dimension represents the  average size of interactions in an input-output mapping.  Calculation of the mean dimension can then help us to understand the complexity of the input-output mapping of an artificial neural network and to compare alternative architectures. Its calculation is pioneered in Owen and Hoyt \cite{OwenHoyt21}.
In order to reduce computational burden, Owen and Hoyt exploit two previous results: Owen's \cite{Owen03} finding that the mean dimension is equal to the sum of the total indices \cite{Owen03} and Jansen's estimator which calculates total indices as variance of one-at-a-time sensitivities \cite{Jansen_cpc_1999}.
Owen and Hoyt then use these results to explore alternative techniques (winding stairs, the naive and the radial approaches) to efficiently sample the required points. 
An additional issue faced in \cite{OwenHoyt21} is that the estimation of the mean dimension requires independent inputs. In order to address the issue Owen and Hoyt consider 13 different joint distributions from which features are independently sampled. A remarkable finding of the experiments in \cite{OwenHoyt21} is that the resulting mean dimension is much lower than the problem dimensionality, independently of the assigned distributions, a finding that would suggest that artificial neural networks look at low order interactions among the features.

The goal of this paper is to continue in the direction of research proposed by \cite{OwenHoyt21}, addressing some of the issues opened in that work. First, we propose a way to compute the mean dimension of an an artificial artificial neural network trained on a given dataset without resampling from a hypothesized distribution. We then perform experiments to study how the mean dimension propagates across the layers of an artificial neural network and how it evolves during training. We use the mean dimension to assess the impact of alternative activation functions, including an experiment with randomly assigned weights. 

We address the problem of feature dependence by first investigating what sensitivity measures Jensen's design yields when features are dependent. We show that one obtains a particular total index, that we denote with $\tau^\prime$, which is a variant of total order indices studied in previous works (see \cite{KuchTara12,MaraTara15}). The work \cite{BorgCapp22} shows that the $\tau^\prime$ indices possess the zero-independence property also when features are correlated. We then study the results yielded by these indices as post-hoc explanations for feature importance. Next we propose the addition of an inverse PCA layer to the artificial neural network \textit{after} training. This addition results in identical outputs but with de-correlated inputs. We then compute the mean dimension of the artificial neural network with respect these inputs. The mean dimension is then employed to compare widely used deep learning architectures like LeNet and ResNet.   

\section{Mean Dimension and Total Indices}

\label{sec:mean dimension}

This section is divided into three parts. Section \ref{sec:maindef} introduces the definitions of total indices and mean dimension. Section \ref{sec:est} presents estimation strategies. Section \ref{sec:dep} discusses the dependent feature case.

\subsection{Main Definitions}
\label{sec:maindef}
The notion of mean dimension is introduced in \cite{CaflMoro97} based on
a central result in statistics, the functional ANOVA expansion. Let $(\Omega
,\mathcal{B}(\Omega ),\mathbb{P})$ be a reference probability space. We
consider features and outputs as a multi-variate random vectors $%
X=(X_{1},X_{2},\dots ,X_{d})$ and $Y=(Y_{1},Y_{2},\dots ,Y_{s})$ on $(\Omega
,\mathcal{B}(\Omega ),\mathbb{P})$ with supports $\mathcal{X}=\mathcal{X}%
_{1}\times \mathcal{X}_{2}\times \dots \times \mathcal{X}_{d}$ and $\mathcal{%
Y}$, and with cumulative distribution functions $F_{X}(x)$ and $F_{Y}(y)$,
respectively (in the reminder, we set the dimensionality of the output equal
to unity for simplicity). We also regard $X$ and $Y$ as linked by a
predictive model, $g:R^{d}\rightarrow R$, with $Y=g(X)$.

Assuming that $g$ is square integrable, and that $F_{X}(x)$ is a product
measure, \cite{EfroStei81} prove the classical functional ANOVA
representation \cite{Sobol1993, Rabitz1999}: 
\begin{equation}
g(x)=\sum_{u\subseteq \{1,2,...,d\}}g_{u}(x_{u}),
\end{equation}%
where the component functions $g_{u}(x_{u})$ are called functional ANOVA
effects and are computed from 
\begin{equation}
g_{u}(x_{u})=\int_{X_{\sim u}}g(x)dF_{{\sim u}}(x_{\sim u})-\sum_{v\subset
u}g_{v}(x_{v}),  \label{fct_Anova}
\end{equation}%
where the symbol $\sim u$ represents the set of all indices with exclusion
of the ones in $u$. We set $g_{\emptyset }=E[g(X)]$. Correspondingly, $F_{u}$
is the cumulative distribution function of the inputs with indices in $u$
and $F_{\sim u}$ is the cumulative distribution function of all inputs but
the ones in $u$. It is possible to prove that the component functions are
orthogonal, that is, given two subset of indices $u$ and $v$, $\int
g_{u}(x_{u})g_{v}(x_{v})dx=0$ as long as $u\neq v$, and that their mean
value is null, i.e., $\int g_{u}dx_{u}=0$. One can then decompose the
variance of $g(X)$, $\sigma ^{2}$, in $2^{n}-1$ terms, writing: 
\begin{equation}
\sigma ^{2}=\sum_{u\subseteq \{1,2,...,d\}}\sigma _{u}^{2}\text{,}
\label{variance_repr}
\end{equation}%
where 
\begin{equation}
\sigma _{u}^{2}=\int_{X_{u}}\left[ g_{u}(\mathbf{x}_{u})\right] ^{2}dF_{u}(%
\mathbf{x}_{u})\text{.}  \label{eq:Vz}
\end{equation}%
The effects $\sigma _{u}^2$ in Equation \ref{eq:Vz} are directly related to
the variance of the corresponding effect functions, and capture the
contribution of the residual interactions among the features whose indices
are in $u$ to the overall variance $\sigma _{Y}^{2}$.

Notice that the first order terms of the variance decomposition can be
written as 
\begin{equation}
\sigma _{i}^{2}=\mathbb{V}[\mathbb{E}[g(X)|X_{i}]]=\sigma _{Y}^{2}-\mathbb{E}%
[\mathbb{V}[g(X)|X_{i}]]\text{,}  \label{eq:si}
\end{equation}%
and thus possess the interpretation of the expected reduction in model
output variance after we are able to fix $X_{i}$. Also, one defines the total indices as%
\begin{equation}
\tau _{i}^{2}=\mathbb{V}[\mathbb{E}[g(X)|X_{-i}]]=\sigma _{Y}^{2}-\mathbb{E}[%
\mathbb{V}[g(X)|X_{-i}]]\text{,}  \label{eq:tauVariance}
\end{equation}%
which is the remaining portion of the variance that is left we fix all
inputs but $X_{i}$. 
In \cite{Homma1996} it is proven that, under independence,%
\begin{equation}
\tau _{i}^{2}=\sum_{v: i \in v} \sigma _{v}^{2}\text{.}
\label{eq:Owentotal}
\end{equation}%
The last equation says that the total index $\tau _{i}^{2}$
coincides with the overall fraction of the variance of the target
contributed by $X_{i}$. Also, we register the notable property that $\tau
_{i}^{2}$ is null if and only if Y does not functionally depend on $X_{i}$.
This is called zero-independence property in \cite{Chat20}, and coincides
with Renyi's postulate D of measures of statistical dependence \cite%
{Renyi1959}.

Dividing the terms of the variance decomposition by $\sigma _{Y}^{2}$, one
obtains the Sobol'variance-based sensitivity indices:%
\begin{equation}
S_{u}=\frac{\sigma _{u}^{2}}{\sigma _{Y}^{2}}\text{.}
\end{equation}%
Noting that $S_{u}\geq 0$ and that $\sum_{u\subseteq \{1,2,...,d\}}S_{u}=1$, 
\cite{CaflMoro97} and \cite{Owen03} interpret the set of $S_{u}$ as an
auxiliary probability mass function over all subsets $u$ of indices. 

Then, the mean dimension is defined as the expectation of the cardinality of these
subsets, \cite{CaflMoro97} \cite{Owen03}). Formally, letting $|u|$
denote the cardinality of $u$, we can write 
\begin{equation}
D_{g}=\sum \limits_{u\subseteq \{1,2,...,d\}}|u|\frac{\sigma _{u}^{2}}{\sigma
_{Y}^{2}}\text{.}  \label{eq:def_mean_dim}
\end{equation}%
Then, $D_{g}$ provides information about the average size of interactions.

\subsection{Estimation Strategies}\label{sec:est}

Using a brute-force estimation strategy for the above quantity would lead to an unacceptably high computational cost.  However, two results make it possible to reduce the computational burden (see \cite{OwenHoyt21}). The first is the relationship between $D_{g}$ and $\tau _{i}^{2}$, proven by Owen (2003) \cite{Owen03}: 
\begin{equation}\label{eq:Dgsumtaui}
D_{g}=\frac{\sum \limits_{i=1}^{d}\tau _{i}^{2}}{\sigma _{Y}^{2}}\text{,}
\end{equation}%
which implies that an estimate of $D_g$ is given by the sum of all total indices.

The second is the finite difference estimator proven in Jansen 1999
\cite{Jansen_cpc_1999}. Let $X^{0}=\{X_{1}^{0},X_{2}^{0},\dots
,X_{i}^{0},\dots ,X_{d}^{0}\}$ and $X^{1}=\{X_{1}^{0},X_{2}^{0},\dots
,X_{i}^{1},...,X_{d}^{0}\}$ be two points in the input space that differ
only in the $i^{th}$ component (henceforth, $d$ denotes the number of features). Then, we call random finite difference the quantity $\Phi
_{i}(X_{i}^{1})=g(X^{1})-g(X^{0})$ and consider the evaluation of this
finite difference at randomized locations in the feature space.  Jensen \cite{Jansen_cpc_1999} shows that $\Phi
_{i}(X_{i}^{1})$ and the total index of $X_{i}$ are related via 
\begin{equation}
\tau _{i}^{2}=\frac{\mathbb{E}[\Phi _{i}^{2}(X_{i}^{1})]}{2}=\dfrac{1}{2}\int_{\mathcal{X%
}}\int_{\mathcal{X}_{i}}\left( g(x_{i}^{1}:x_{-i})-g(x)\right)
^{2}dF_{X_{i}}(x_{i}^{1})\prod\limits_{s=1}^{d}dF_{X_{s}}(x_{s})\text{.}
\label{eq:tauVphi}
\end{equation}
Then, by \eqref{eq:Dgsumtaui}, we have
\begin{equation}
D_{g}=\frac{\sum_{i=1}^{d}\mathbb{E}[\Phi^2_{i}(X_{i}^{1})]}{2\sigma _{Y}^{2}}%
\text{.}  \label{eq:tauDg}
\end{equation}%
Owen and Hoyt exploit \eqref{eq:tauDg} for devising an estimation strategy that
compares evaluations of the predictions of an artificial neural network at $X^{0}$ and $%
X^{1}$. The corresponding estimator has a cost of the order of $d(N+2)$ model evaluations. One key aspect in this procedure is the selection of the
replicate $X_{i}^{1}$. This can be executed in alternative ways. In \cite{OwenHoyt21}, three methods are compared: naive, radial and winding stair. While we
do not enter into the details of each procedure, we notice a common aspect:
the new value $X_{i}^{1}$ is sampled independently of $X_{-i}^{0}$. While this is not a problem when the features are independent,
the situation changes in the case of dependent features.

\subsection{Dependent Features}
\label{sec:dep}

When features are dependent, one of
the main assumptions that underlies Equations \eqref{variance_repr}, (\ref{eq:Owentotal}) and (\ref{eq:def_mean_dim}) does not hold anymore. We refer
to works such as \cite{Hook07}, \cite{Chastaings2012}, \cite{Rahm14}, %
\cite{Li2011GeneralFO},\cite{BorgMorr18} and \cite{Rahm18} for an overview of theoretical
aspects. Closely related to our work are the observations that Equations %
\eqref{eq:si} and \eqref{eq:tauVariance} still hold. In this respect, the
interpretation of both the first and total order indices in terms of
variance-reduction remains well posed also when inputs are dependent. It is
also worth recalling that total indices can still be written in terms of
finite differences (see \cite{KuchTara12}, \cite{MaraTara15}, \cite{GeMene17}, \cite{MaraBeck21} among others) as
\begin{equation}
\tau _{i}^{2}=\frac{1}{2}\int_{\mathcal{X}_{i}}\int_{\mathcal{X}}\left(
g\left( x_{i}^{1}:x_{-i}\right) -g\left( x\right) \right) ^{2}dF_{X_{i}|%
\mathbf{X}_{-i}}(x_{i}^{1}|x_{-i})dF_{\mathbf{X}}(x)\text{,}
\label{eq:tauicorr}
\end{equation}%
where $F_{X_{i}|\mathbf{X}_{-i}}(x_{i}^{\prime }|\mathbf{x}_{-i})$ is the
conditional distribution of $X_{i}$ given $\mathbf{X}_{-i}$. Equation %
\eqref{eq:tauicorr} suggests that the new point $x_{i}^{1}$ is sampled
conditionally on the values of the remaining inputs, an operation that
can be performed in a given-data context. We observe that under input
dependence $\tau _{i}^{2}$ loses the zero-independence property
(see \cite{KuchTara12} among others). 

If we still sample $X_{i}$ independently we obtain
\begin{equation}
\label{eq:tauiprime}
\tau _{i}^{\prime}=\frac{1}{2}\int_{\mathcal{X}%
_{i}}\int_{\mathcal{X}}\left( g\left( x_{i}^{\prime}:x_{-i}\right) -g\left( x\right) \right)
^{2}dF_{X_{i}}(x_{i}^{\prime})dF_{\mathbf{X}}(x)\text{.}
\end{equation}%
In \cite{BorgCapp22} it is shown that $\tau _{i}^{\prime}$ is null if an only if the target or forecast is independent of $X_i$. Thus, while $\tau _{i}^{\prime}$ differs from $\tau _{i}^{2}$
under dependence, it still enjoys the zero-independence property.

\section{Estimating the Mean Dimension from Available Data
\label{sec:estimating_md}}

Two aspects of the previous discussion raise potential challenges in computing the mean dimension: we need to re-sample the
features ($X_{i}^{^{\prime }}$) and we need to take correlations into
account. We consider these issues in the next two subsections, respectively.

\subsection{Using Given-Data to Find Random Finite Differences}
\label{sec:featresampl}

An intuition to avoid sampling from assigned distributions is the following. Suppose we have a labeled dataset of inputs $x$ and an artificial neural network
trained on this data. Let $x^{k}=(x_{1}^{k},...,x_{d}^{k})$ denote the $%
k^{th}$ input with $k=1,2,\dots ,N$, where $N$ is the sample size. Let also $%
y^{k}=g(x^{k})$ be the network prediction corresponding to input $x^{k}$. We
then consider modified samples where a single feature value (say $x_i^k$) has been
replaced by a value of the same feature but from a different and randomly picked realization (say $x_i^l$), with $l\neq k$. We denote such
modified inputs as 
\begin{equation}
(x_{i}^{l};x_{\sim i}^{k})=(x_{1}^{k},\ldots
,x_{i-1}^{k},x_{i}^{l},x_{i+1}^{k},\ldots ,x_{d}^{k}),  \label{new_pair}
\end{equation}%
where the sequence $(x_{i}^{l};x_{\sim i}^{k})$ is created by replacing $%
x_{i}^{k}$ with $x_{i}^{l}$ in $x^{k}$. The finite change with respect to
this modification is then defined as 
\begin{equation}
\phi _{i}^{kl}=g(x_{i}^{l};x_{\sim i}^{k})-g(x^{k}).  \label{finite_changes}
\end{equation}%
These finite changes are computed for $r$ pairs of inputs, $x^{k}$ and $%
x^{l} $, uniformly sampled from the dataset. This yields an $r\times d$%
-matrix of finite differences, with $r$ finite changes for each input
feature.

Using the plug-in principle, the values of $g(x_{i}^{l};x_{\sim i}^{k})$ and 
$g(x^{k})$ yields estimates of the total indices of the type%
\begin{equation}
\widehat{\tau }_{i}^{2}=\sum_{k=1}^{r}\frac{(g(x_{i}^{l};x_{\sim
i}^{k})-g(x^{k}))^{2}}{2(r-1)}\text{.}  
\label{eq:hatphi}
\end{equation}
Then, by Equation \eqref{eq:tauDg}, we obtain the following estimate of the
mean dimension 
\begin{equation}
\widehat{D}_{g}=\sum_{i=1}^{d}\sum_{k=1}^{r}\frac{(g(x_{i}^{l};x_{\sim
i}^{k})-g(x^{k}))^{2}}{2(r-1)\widehat{\sigma }_{Y}^{2}}\text{,}
\label{eq:hatDg}
\end{equation}%
where $\widehat{\sigma }_{Y}^{2}$ is an estimate of the variance of the
artificial neural network forecasts. Note that, summing over all possible values of $%
x_{i}^{l}$, $l=1,2,...,r$, one can also define the following $U$-statistic
as an estimate for $\overline{\tau }_{i}^{2}:$ 
\begin{equation}
\widehat{\tau }_{i}^{U}=\sum_{l=1}^{r}\sum_{k=1,k\neq l}^{r}\frac{%
(g(x_{i}^{l};x_{\sim i}^{k})-g(x^{k}))^{2}}{2r(r-1)}\text{.}
\label{eq:tauiU}
\end{equation}%
Specifically, $\widehat{\tau}_{i}^{U}$ is a U-statistic of degree 2. Then,
we have the following estimate of the mean dimension:%
\begin{equation}
\widehat{D}_{g}^{U}=\sum_{i=1}^{d}\sum_{l=1}^{r}\sum_{k=1,k\neq l}^{r}\frac{%
(g(x_{i}^{l};x_{\sim i}^{k})-g(x^{k}))^{2}}{2r(r-1)}\text{.}  
\label{eq:DgU}
\end{equation}
We note that $\widehat{D}_{g}^{U}$ is a U-statistic. In fact, the sum of any
two terms in $\widehat{D}_{g}^{U}$ is a $U$-statistic (see 
\cite{BoseChat18}). This means that the estimates in \eqref{eq:tauiU} and 
\eqref{eq:DgU}  are asymptotically normal and unbiased. 

Under feature independence, $\widehat{\tau }_{i}^{U}$ and $\widehat{D}%
_{g}^{U}$ are estimates of $\tau_{i}^{2}$ and $D_{g}$, respectively. However, under feature dependence, $\widehat{\tau}_{i}^{U}$ results in an estimate of $\tau_{i}^{\prime }$ in
\eqref{eq:tauiprime}, while $\widehat{D}_{g}^{U}$ loses its interpretation as mean dimension. An empirical approach for still using the design and calculating the mean dimension for artificial neural networks on uncorrelated features is discussed in the next section.

\subsection{Inverse PCA Layer}
\label{sec:inverse_pca_layer}
\begin{figure}
    \centering
    \includegraphics[width=\textwidth]{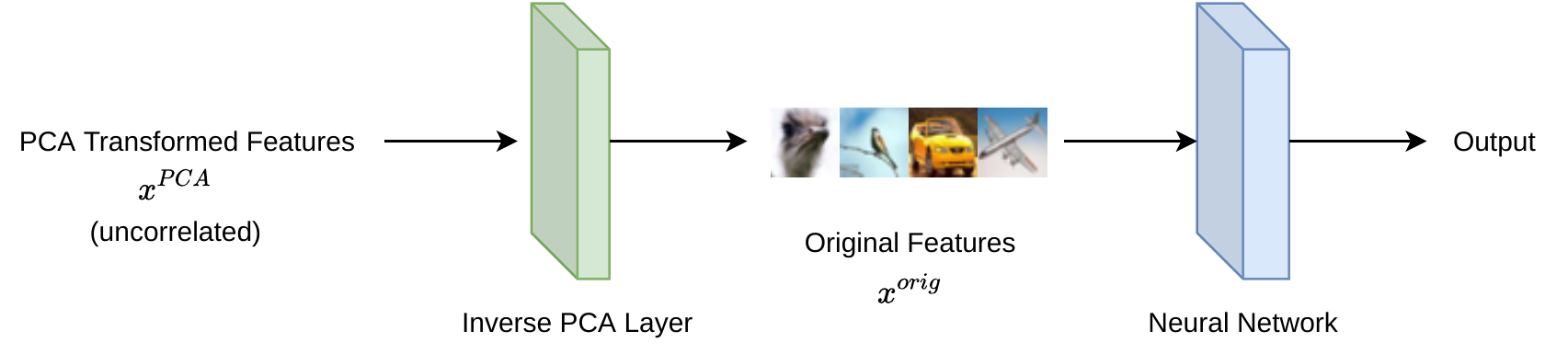}
    \caption{Artificial neural network architecture with inverse PCA layer. The artificial neural network is trained on the original features $x^{org}$. After training has ended, the inverse PCA layer is added, which takes PCA transformed features $x^{PCA}$ as inputs and transforms them back to the original features $x^{org}$. The extended network is then used for estimating the mean dimension.}
    \label{fig:pca_diagram}
\end{figure}

We first note that applying a PCA transformation to the original features $x^{org}$ leads to transformed features $x^{PCA}$ which are (linearly) uncorrelated, and therefore one possible approach is to train the artificial neural network on PCA transformed features instead of the original features. This, however, would lead to a different network and possibly distort the results. We therefore adopt a different approach and extend the artificial neural network  \textit{after} training on the original data by inserting an inverse PCA layer at the beginning of the network, see Fig~\ref{fig:pca_diagram}. The inverse PCA layer transforms PCA transformed features $x^{PCA}$ back to the original features $x^{orig}$. The final output of the extended network on PCA transformed data $x^{PCA}$ is therefore identical to the output of the original network on the original data $x^{org}$. We then use the extended network with uncorrelated PCA transformed features $x^{PCA}$ as inputs for estimating the mean dimension. That is, we apply the Jensen's design using one-at-a-time variations of $x^{PCA}$ in Equation \eqref{finite_changes}. Then, the finite change on $x^{PCA}$ is transferred to $x^{\text{orig}}$ by the inverse PCA layer and the output of the artificial neural network is recorded. 

While the inverse PCA layer does not lead to strictly independent features and calculates the mean dimension in the space of PCA features $x^{PCA}$ and not the original features $x^{org}$, we show in the following sections that it can still be a viable tool for comparing network architectures and monitor changes in the network during training. 

\section{Artificial Neural Networks as Surrogates: Experiments with Simulated Data}
\label{sec:experiments}

Recently, Taverniers et al. \cite{TaveTart21} study the interpretability of artificial neural networks employed as data-driven surrogate models within the simulation of complex, multiscale systems. In this case, the artificial neural network is trained on a synthetic dataset and used to replace parts of the modeling process that are time consuming.
The difference between a purely data driven application in this case is that the data are generated from a given, albeit not explicitly known, input-output mapping. In this case, the theory of artificial neural networks and in particular the universal approximation result of \cite{Hornik89} suggests us an additional interpretation. Let $\hat{g}(\mathbf{x})$ and $g(\mathbf{x})$ denote the network input-output mapping and  the original input-output mapping, respectively. Then, \cite{Hornik89}'s approximation theorem suggests that, in the case of an accurate approximation, the artificial neural network has learned the true input-output mapping structure and we then expect that the estimated $D_{\hat{g}}$ not only represents the mean dimension of the artificial neural network, but is also informative on the mean dimension of the true input-output mapping. 

In this section, we propose a series of  experiments on well known test cases of small dimensionality to illustrate key concepts and some inferences that can be made with the mean dimension. In Section \ref{sec:ishigami} we provide results of several experiments performed on the Ishigami function. In Section \ref{sec:levele}, we presents results for the Level E code. 

\subsection{An Analytical Test Case: Ishigami} \label{sec:ishigami}
The first experiments are centered around a regression problem with a scalar output. We use a dataset generated from the well known Ishigami function \cite{151285}, which implements the function
$$g(x_1,x_2,x_3)= \sin(x_1)+ 7\:\sin^2(x_2)+0.1\: x_3^4 \sin(x_1).$$
The features $x_j$ are uniformly and independently distributed between $-\pi$ and $\pi$ for $j=1,2,3$.
This function is often used as a test for uncertainty and sensitivity analysis methods.

The mean dimension is known analytically for this model \cite{Kucherenko2015} and is equal to
$ D_{\text{f}}=1.24 $. 

We use PyTorch \cite{paszke2019pytorch} for training the artificial neural network. We generate a dataset consisting of 60,000 instances. One instance is composed of the randomly drawn features and the Ishigami function evaluation at these inputs as corresponding output. The data is then split into $48,000$ samples for training and $12,000$ samples for testing. The network architecture is depicted in Fig. $\ref{results_table}$. We use $300$ nodes in the first layer and $50$ nodes in the second one. We train for $100$ epochs with the learning rate set to $0.01$. Weights and biases are initialized using the Pytorch default Kaiming initializiation \cite{7410480}. We report results for experiments with both the Rectified Linear Unit (ReLU) and the hyperbolic tangent (TanH) activation functions.

The estimation routine from Section \ref{sec:estimating_md} yields a very precise estimate of the mean dimension for any of the used activation functions. Using a mean of $20$ replicates, we obtain an average mean dimension of $1.241$ with a standard deviation of $6.6 \times 10^{-3}$ using the Rectified Linear Unit (ReLU) activation function.
Thus, in terms of the mean dimension, the learned artificial neural network is a very accurate meta-model of the Ishigami function.  

\begin{figure}[t]
\centering
\includegraphics[width=0.8\textwidth]{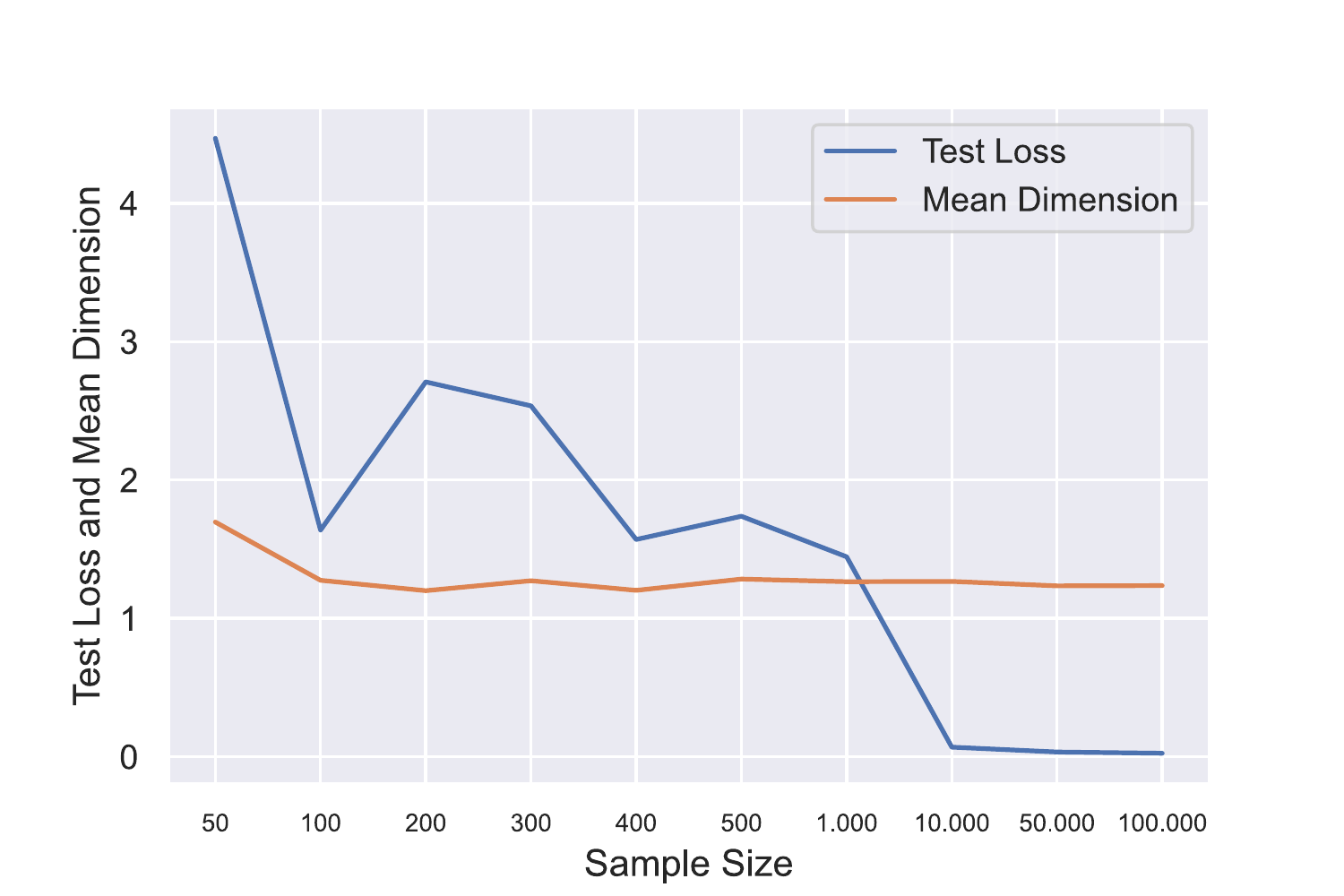}
\caption{\label{table_conv} Mean dimension estimate (MD) and mean square error (loss) as the sample size increases for a small feed-forward artificial neural network with ReLU activation function, trained on the Ishigami function. Note that the number of finite differences used for estimating the mean dimension scales linearly with the sample size.}
\label{fig:convergence_md}
\end{figure}

Fig. \ref{fig:convergence_md} shows the estimated mean dimension and mean squared error at increasing sample sizes. 
Note that $N$ here refers to the sample size used to train the model and also to the number of pairs used to estimate the mean dimension ($r$ in Section \ref{sec:estimating_md}), as in our experiments we use $r=N-1$. The mean dimension approaches the value $D_g=1.24$ already at a sample size of about $N=200$. This result is also in agreement with the universal approximation function theorem of Hornik 1989 and is a sign that the network does learn the original input-output mapping. 

We highlight two aspects of these findings. First, we repeated the experiments using the hyperbolic tangent (TanH) activation function, obtaining very similar results. Then, we added noise to the data by augmenting the feature set with 7 dummy variables, for a total input dimension $d=10$.
The additional features are random replicates of the three true inputs, drawn uniformly between $-\pi$ and $\pi$ and independent of the output $y$. 
We keep the original output in the dataset. Retraining the network on these noisy data and estimating the mean dimension leads to results again similar to the ones we just illustrated.

\subsection{Where Does the Mean Dimension Grow?} 
\label{sec:Ishigami_layer}
To study how the mean dimension changes throughout the network, we regard each neuron in the network as an output and calculate the corresponding mean dimension. Fig. \ref{results_table} provides a visualization of the results and a sketch of the architecture.
Each neuron in Fig. \ref{results_table} has a corresponding mean dimension. In the table below the network architecture, however, we report the layer average mean dimension (LAMD). Also, for each layer we consider the mean dimension before and after activation (to illustrate the numbers $1.008$ and $1.1891$ refers to the LAMD of the first layer without and with activation, respectively). We perform this layer-by-layer analysis for an artificial neural network trained on the Ishigami data (Section \ref{sec:NN_on_Ishigami}) and for a random artificial neural network using random data as input (Section \ref{sec:NN_on_randomdata}). 

\subsubsection{Artificial Neural Network Trained on Ishigami Data} \label{sec:NN_on_Ishigami}
\begin{figure}[t]
\centering
\includegraphics[width=0.8\textwidth]{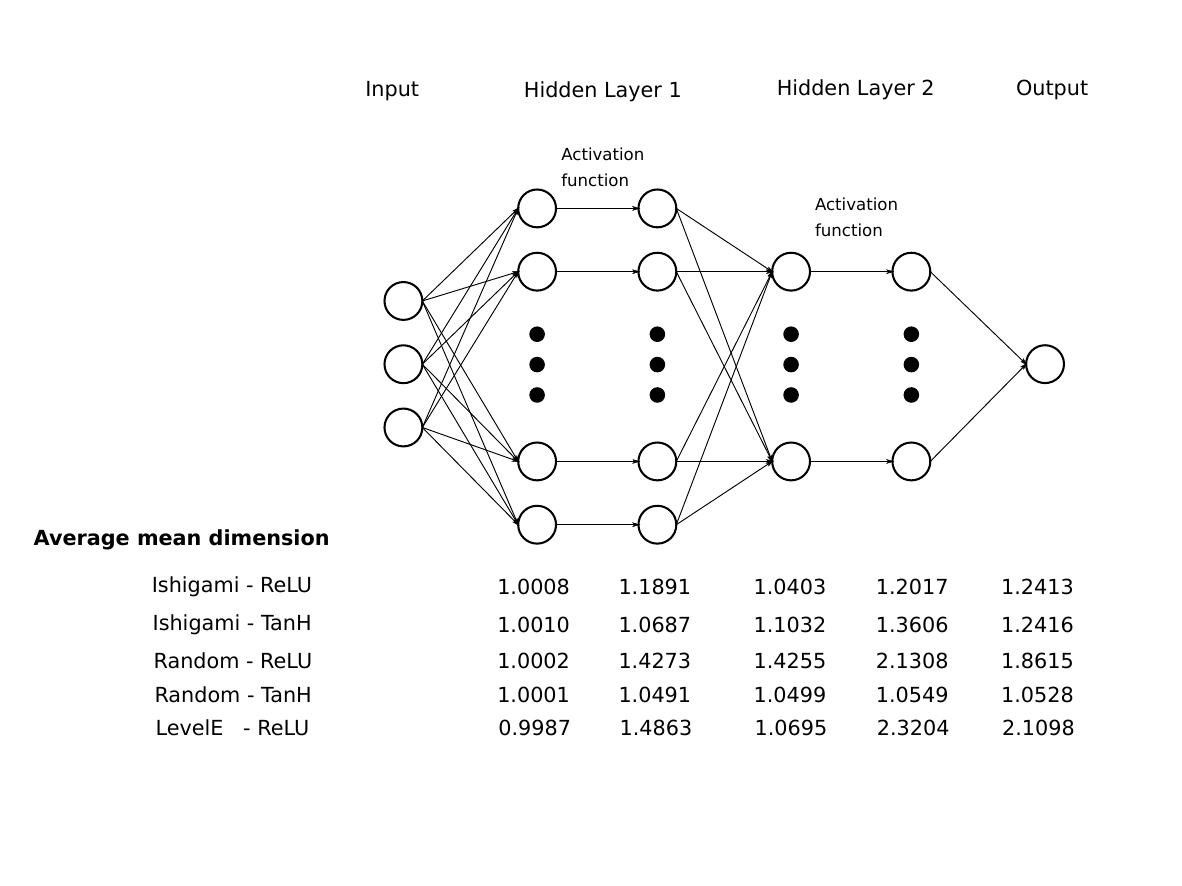}
\caption{Sketch of the artificial neural network architecture used for the Ishigami experiments (Section \ref{sec:ishigami} and \ref{sec:NN_on_Ishigami}) \& the random model and random data experiment (\ref{sec:NN_on_randomdata}). Below: Table with average of $20$ replicates of the per layer average mean dimension for different experiment set ups and different activation function (ReLU and TanH). The layer architecture has the following structure: fully-connected layer, activation function, fully-connected layer, activation function, fully-connected layer.}
\label{results_table}
\end{figure}

The table in Fig. \ref{results_table} shows the average over $20$ replicates of the LAMD. The first two rows show the LAMD values using the Ishigami data for a network with ReLU activation and with TanH activation, respectively. The even columns report results without activation, the odd columns after activation. 

One notes the strong increase in LAMD following an activation. For instance  going from column 1 to column 2 in the first row (with ReLU activation), we observe a systematic increase in the LAMD of the first layer. The same holds for the second application of activation function, (column 3 to column 4). While this is consistent for both activation functions, the ReLU activation gives a higher increase in the LAMD of the first layer, while TanH does so in the LAMD of the second layer. Note that the mean dimension of the second layer on the Ishigami data with TanH activation is higher than the output mean dimension. Because the output mean dimension is the same with both activation functions, the LAMD has then to decrease in order to get to the correct estimate.

Overall, interactions seem to arise due to the non-linear activation functions, as one would expect, while fully-connected layers (before activation) do not lead to increases in the mean dimension.
We also get a mean dimension very close to 1 for each pre-activation neuron in the first layer, which corresponds to a linear transformation of the inputs.

\subsubsection{A Random Artificial Neural Network \& Random Input} \label{sec:NN_on_randomdata}

In these experiments, we aim at further testing the differential effects of the activation function on the mean dimension.  We consider a neural network, with the same structure as the one in Fig. \ref{results_table}.
However, we now consider a problem with increased dimensionality, with $d=200$  i.i.d. standard normal features. We do not train the network but assign a random value to its parameters via the Kaiming initialization of Pytorch \cite{7410480}.
We generate a sample of size $60.000$ and gather the corresponding predictions of the artificial neural network, with either ReLU or TanH activations. The third row of the table in Fig. \ref{results_table} shows that the mean dimension of the network with ReLU activation functions is $1.86$ when averaging over 20 replicates. The average mean dimension of the same experiment using TanH activation function is $1.05$. 
This indicates that even when using random data, the ReLU activation function induces higher average size of interactions than the TanH activation function.  
These results are in line with the experiments performed on the Ishigami function (Section \ref{sec:NN_on_Ishigami}).

\subsection{Level E} 
\label{sec:levele}
The Level E simulator is a computer program that encodes the solution to a set of differential equations used in safety assessment for nuclear waste disposal \cite{saltelli_jasa_2002}.
The equations predict the radiological dose to humans released by the underground
migration of radionuclides from a nuclear waste disposal site over geologic time scales.
The simulation model includes 12 uncertain features, such as the length of the first geosphere layer or the retention factor for the chain elements in the second layer.
The quantity of interest in this study is the maximal radiological dose due to the four radionuclides ($^{129}I$ and the
chain $^{237}Np \rightarrow ^{233}U \rightarrow ^{229}Th$). The dataset consists of 70000 input-output realizations.
The training sample is of size 50000 and we use 20000 instances for testing. We train a two layer artificial neural network with ReLU activation on data generated by the level E code.
The input and output data is normalized, the learning rate set to 0.001 and we stop training after 120 epochs. 
We register a mean square error of 7.124e-05 and 2.548e-05 and an $R^2$ of 0.974. 

We apply the design in illustrated in Section \ref{sec:featresampl}. Because the inputs are independent, the design yields Sobol' total indices whose estimates can be summed to yield the mean dimension. 

Table \ref{tab:levele} reports the estimates of the total indices obtained from the artificial neural network at the end of training. The numerical estimates are in line with results in the literature obtained using the original model \cite{Saltelli1999,saltelli_jasa_2002}, in accordance with Hornik's universal approximation theorem. In terms of feature importance, the first and the fourth inputs are ranked as most relevant ones (also in agreement with previous literature).

\begin{table}[H]
\centering
\begin{tabular}{|c|c|c|}
\hline
\textbf{Input (Symbol)}                                      & $\hat{\tau}^2_i$ & $\hat{\tau}^2_i/\sigma^2_Y$ \\ \hline
Containment time    (T)                                        & 6.81e-06 & 0.00          \\ \hline
Leach rate for iodine ($k_\text{l}$)                                      & 6.33e-06  & 0.00         \\ \hline
Leach rate for Np chain nuclides ($k_\text{C}$)                            & 1.57e-05     &0.01      \\ \hline
Water velocity in geosphere’s 1st layer  ($v^{(1)}$)                  & 1.15e-03          &  0.63  \\ \hline
Length of geosphere’s 1st layer ($\ell^{(1)}$)                            & 2.20e-04  & 0.12         \\ \hline
Retention factor for I (1st layer)  ($R^{(1)}$)                         & 4.91e-05   & 0.03          \\ \hline
Factor to compute ret. coeff. for Np (1st layer) ($R_C^{(1)}$)  & 3.83e-04    &0.21       \\ \hline
Water velocity in geosphere's 2nd layer   ($v^{(2)}$)                  & 2.41e-04  & 0.13         \\ \hline
Length of geosphere’s 2nd layer ($\ell^{(2)}$) 
                             & 1.21e-04 & 0.07         \\ \hline
Retention factor for I (2nd layer) ($R^{(2)}$) 
                         & 3.01e-05    &0.02       \\ \hline
Factor to compute ret. coeff. for Np (2nd layer) ($R_C^{(2)}$)  & 2.40e-04     & 0.13     \\ \hline
Stream flow rate      (W)                                      & 1.19e-03        & 0.66   \\ \hline
\end{tabular}
\caption{Inputs of the level E computer code and corresponding estimate of $\tau^2_i$.}
\label{tab:levele}
\end{table}

\begin{figure}[H]
    \centering
    \includegraphics[width=0.5\textwidth]{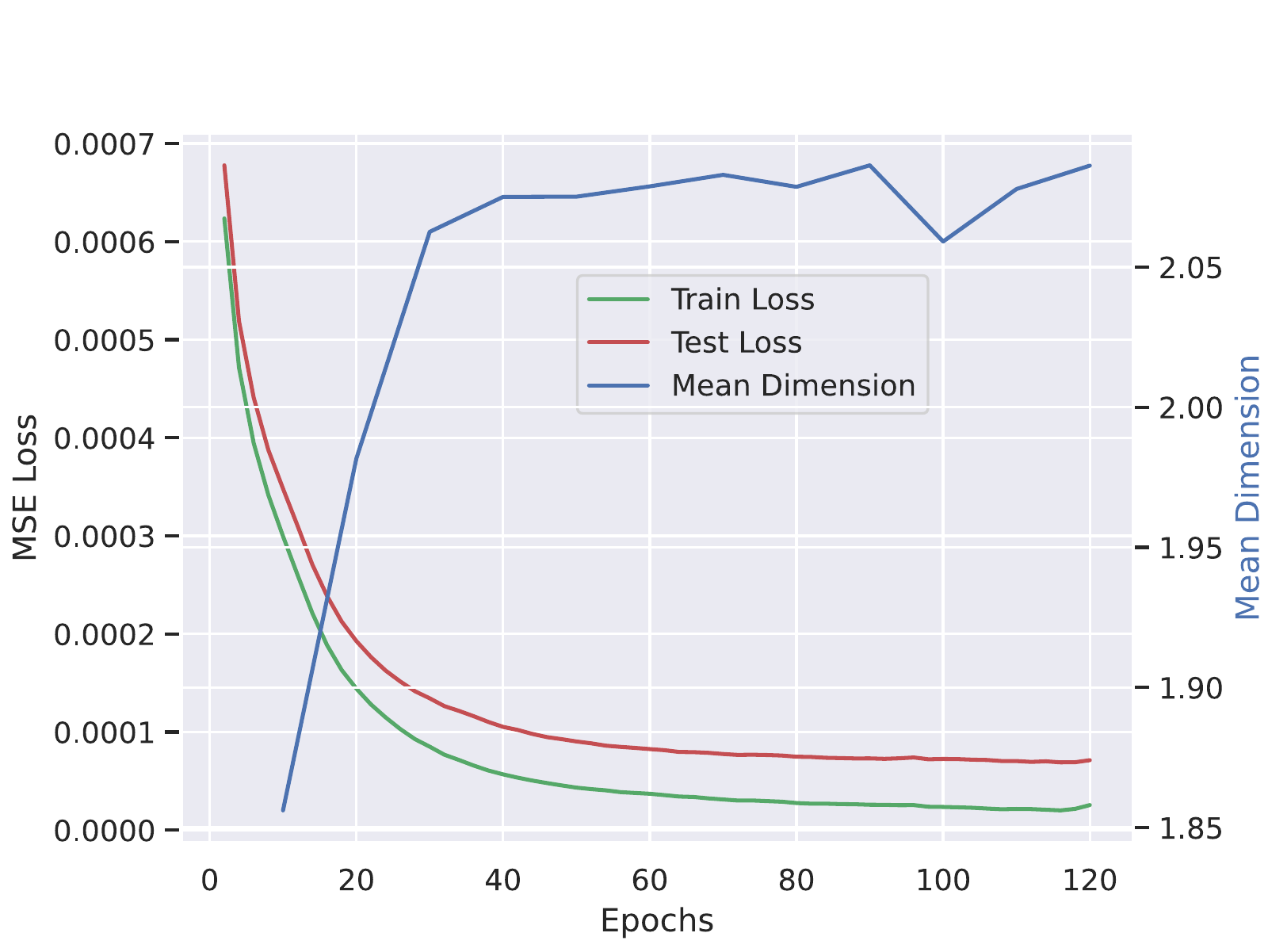}
    \caption{Level E: Mean dimension (MD, blue), training (green) and test loss (red) evolution during training of 120 epochs. The mean dimension is estimated every $10^{th}$ epoch of training. Errors are the mean square error (MSE) in the training and test data.}
    \label{fig:levele}
\end{figure}

The values of the total indices yield a mean dimension $\hat{D}_g\approx 2.10$. Previous studies have shown that the Level E input-output mapping is characterized by strong interactions. This finding confirms that the artificial neural network is capable of well approximating such complex input-output relationship.

The evolution of the mean dimension during training is shown in Fig. \ref{fig:levele}.
We observe that $D_g$ increases initially and stabilizes after 40. Notice that after this number of epochs the train and test losses also reach a plateau. 
Finally, we report the layer-by-layer LAMD of the neural network in the last row of the table in Fig. \ref{results_table}. The values are in line with previous experiments, and display a mean dimension increasing as layers gets deeper, with more relevant jumps occurring under a ReLU activation function.

\section{Image Classification: The Case of Dependent Inputs}\label{sec:real_world}
We consider an image classification task with a well known dataset and apply the methodology explained above to compare alternative artificial neural network architectures trained on the same dataset. To allow reproducibility of the experiments, we select the well known benchmark dataset Cifar10 \cite{Krizhevsky2009}.
Cifar10 contains $60,000$ labeled images of objects belonging to 10 different classes. The images consist of $32 \times 32$ pixels with $3$ color channels. 

Three network architectures that have been previously explored in literature and tested on Cifar10 are Lenet-5 \cite{lecun2015lenet}, the DenseNet-121 \cite{Huang2017} and ResNet-101 \cite{He2016}.
All three are convolutional artificial neural networks, but they differ in their complexity and layer types. 
The LeNet-5 architecture we use consists of an input layer, two convolutional layers with max-pooling and two linear layers. 
The ResNet and DenseNet architectures we employ are deeper and more complex. ResNet architectures are characterized by their use of residual connections, which connect non-consecutive layers, effectively skipping parts of the network in the flow of information \cite{He2016}. DenseNets follow a similar idea and connect hidden layers with all preceding hidden layers instead of a single preceding one \cite{Huang2017}. Both ResNets and DenseNets can be trained in very deep versions, with ResNet-101 consisting of $101$ layers and DenseNet-121 of $121$ layers. We refer to the original publications for a detailed description of the single blocks of these architectures. We transform the output of all networks to class probabilities by a softmax operation and use the cross-entropy loss as a training objective.

Two issues arise when trying to estimate the mean dimension following the routine in Section \ref{sec:estimating_md}. First, the estimation procedure would require independent features to yield the mean dimension. Second, estimation of the mean dimension requires a scalar output, while a classification problem usually yields a set of probabilities, one for each class.
We address the the first issue adding the inverse PCA layer described in Section \ref{sec:inverse_pca_layer} to the artificial neural network. For the second issue, we by start noting that the final layer of artificial neural networks in classification is usually a softmax layer, which yields a number of outputs equal to the number of classes.
The resulting probabilities are natural outputs of interest: They need to be sensitive to features in the input that relate to the classes to be predicted and are typically the direct input to the objective function that is used to train the network. We could analyze the single outputs independently, as is done in \cite{OwenHoyt21}, and also in the layer-by-layer analysis we have performed in the previous sections. However, in this section, we would like to have a single number characterizing the mean dimension of the artificial neural network. When using the cross-entropy for training the artificial neural network, the objective function for a single input-output pair is the negative log-probability of the true class for the input. This quantity depends on all neurons in the last layer before the softmax and is the value being minimized during training. We therefore set the output $Y$ equal to the negative log-probability of the true class in the remainder of this analysis.

Except where specified otherwise, all artificial neural networks in this section are trained for $120$ epochs, with a learning rate of $0.001$ using the $Adam$ optimizer \cite{Kingma2015}. The weights and biases are initialized by the default Kaiming initialization in Pytorch \cite{7410480}.

We then apply the designs of Section \ref{sec:featresampl} and \ref{sec:inverse_pca_layer} to obtain insights on the mean dimension and on the feature importance.

\subsection{$\tau^\prime_i$ indices as Explanations} 
First, we consider the effects $\phi _{i}^{kl}$ in \eqref{finite_changes} evaluated as differences between $Y$ computed for image $k$, and $Y$ for the image obtained by substituting the value of pixel $X_i$ in image $k$, $x_i^k$, with the value of  pixel $X_i$ in another randomly picked image, $x_i^l$. We estimate $20000$ finite changes and, from these, we obtain estimates of $\tau_i^\prime$.

We visualize the $\hat{\tau}_i^\prime$ estimates as heatmaps in two different ways: 
\begin{enumerate}
    \item We provide an individual heatmap for each channel (red, green, blue), called channel heatmap, henceforth.
    \item We provide a final heatmap, which aggregates explanations of the three color channels at the same position, by taking the mean of the channel importance values (aggregated heatmap, henceforth).
\end{enumerate}
\begin{figure}[H]
\centering
    \begin{subfigure}[b]{1\linewidth}
    \centering
    \includegraphics[scale=0.5]{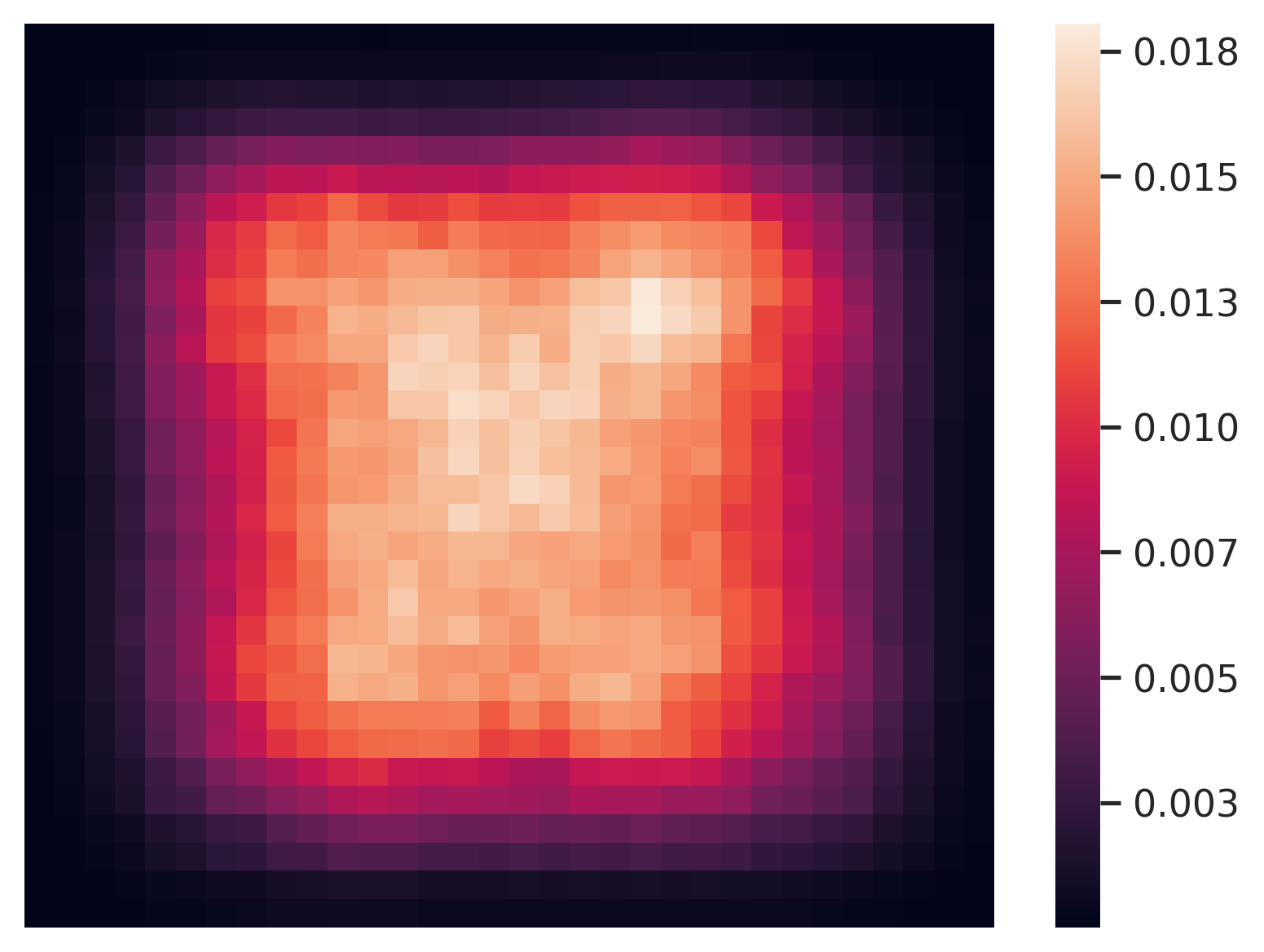}
    \end{subfigure}
    \begin{subfigure}[b]{1\linewidth}
    \centering
    \includegraphics[scale=0.5]{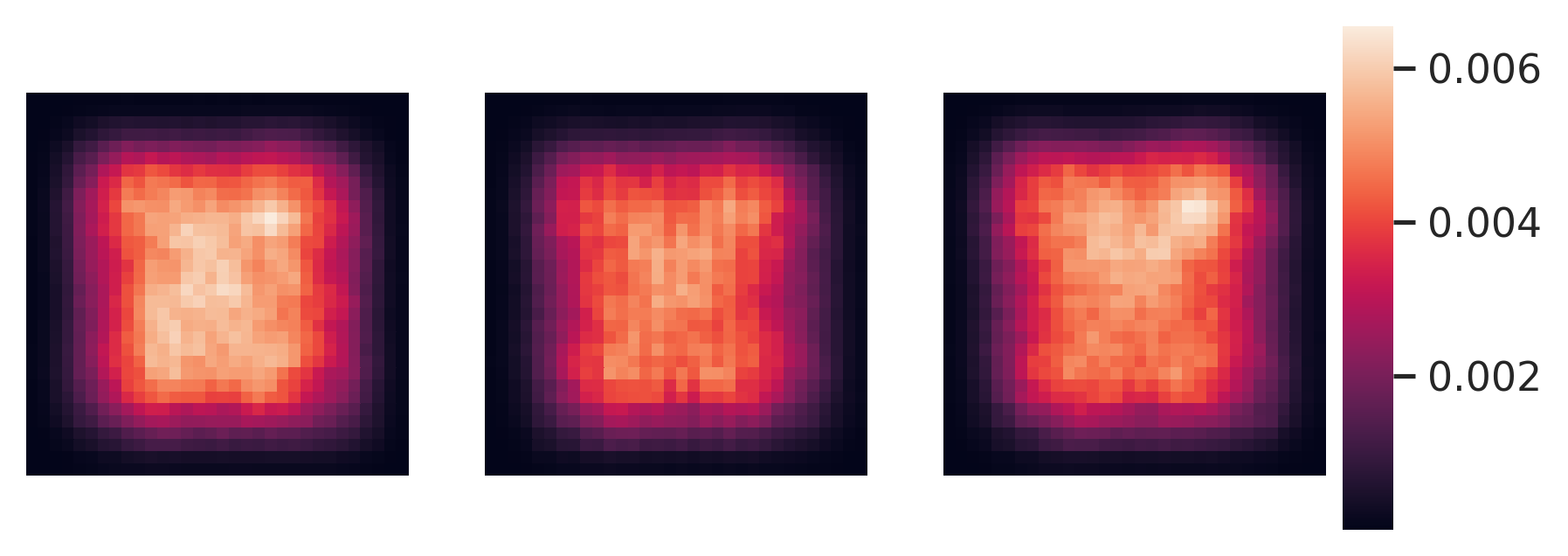}
    \end{subfigure}
    \caption{Heatmaps of $\hat{\tau}_i^\prime$ for the \textbf{Lenet} architecture. Above: Aggregated heatmap. Below: Channel heatmaps}
    \label{fig:heatmap_lenet}
\end{figure}
\begin{figure}[H]
\centering
    \begin{subfigure}[b]{1\linewidth}
    \centering
    \includegraphics[scale=0.5]{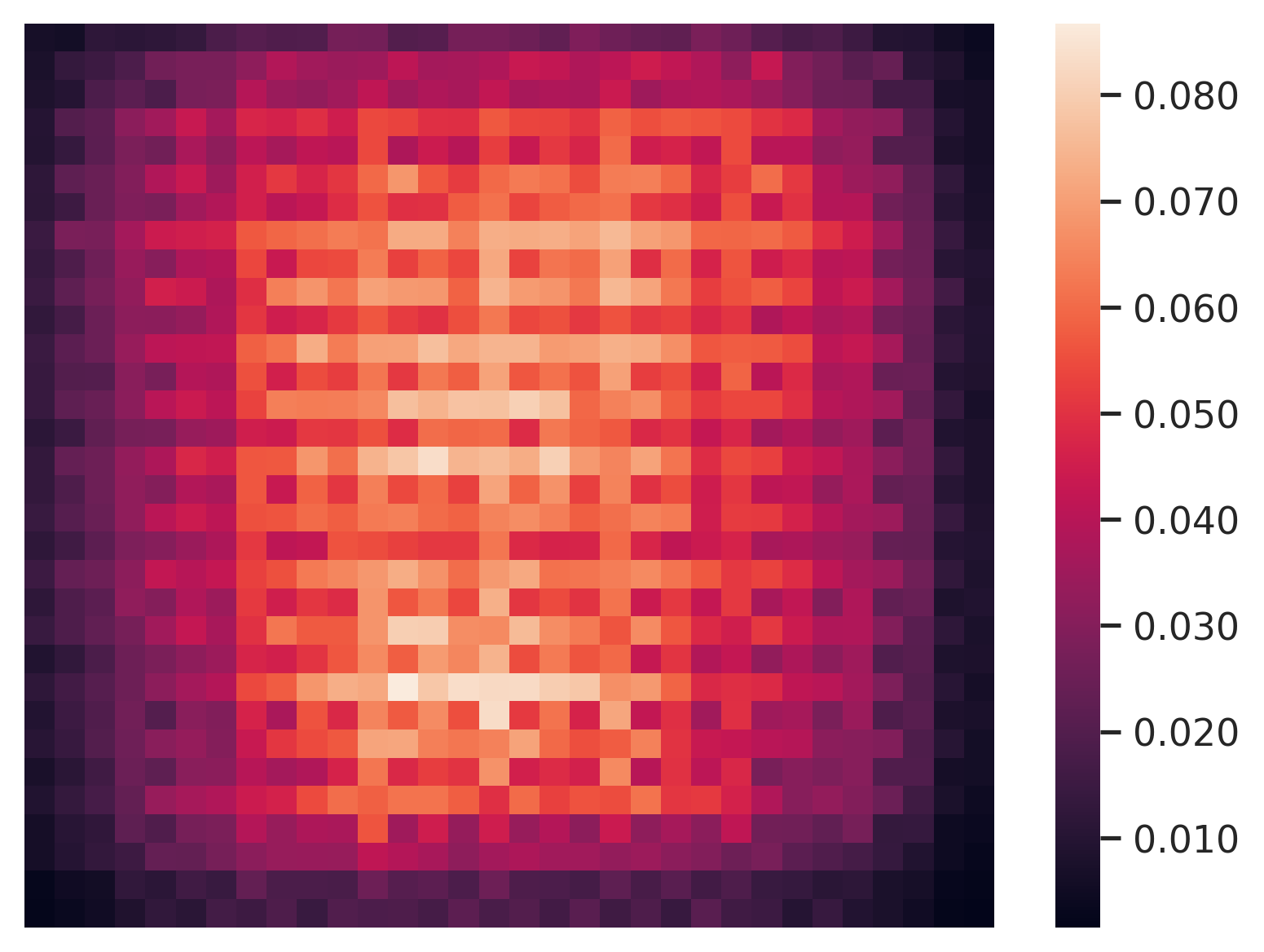}
    \end{subfigure}
    
    \begin{subfigure}[b]{1\linewidth}
    \centering
    \includegraphics[scale=0.5]{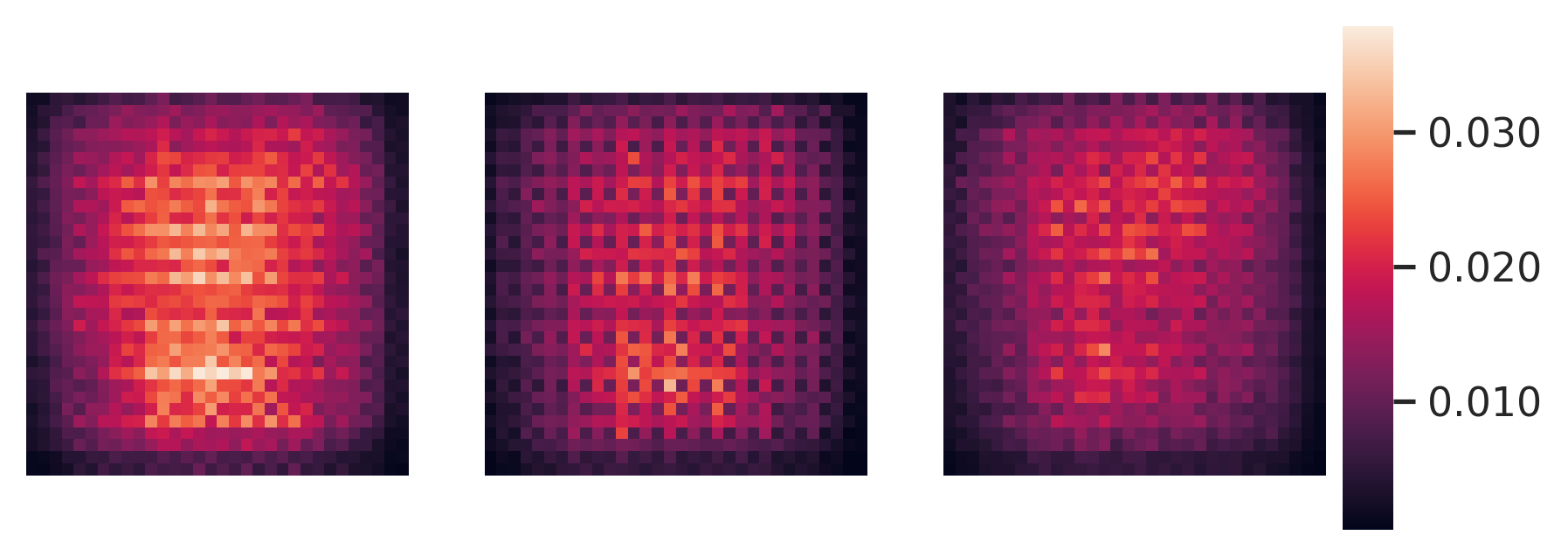}
    \end{subfigure}
    
    \caption{Heatmaps of $\hat{\tau}_i^\prime$ for the \textbf{Densenet121} architecture. Above: aggregated heatmap. Below: channel heatmaps.}
    \label{fig:heatmap_densenet}
\end{figure}
\begin{figure}[H]
\centering
    \begin{subfigure}[b]{1\linewidth}
    \centering
    \includegraphics[scale=0.5]{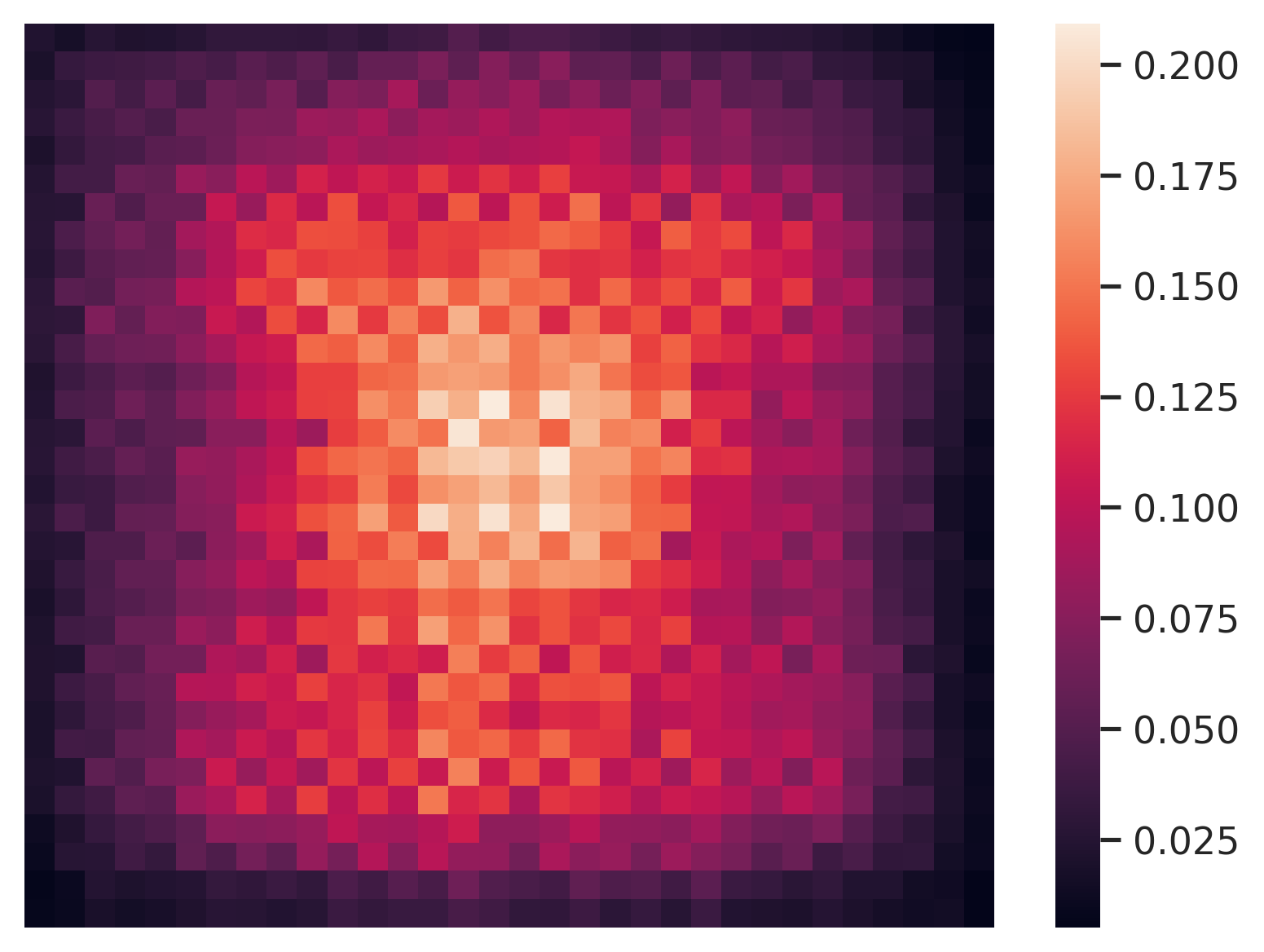}
    \end{subfigure}
    
    \begin{subfigure}[b]{1\linewidth}
    \centering
    \includegraphics[scale=0.5]{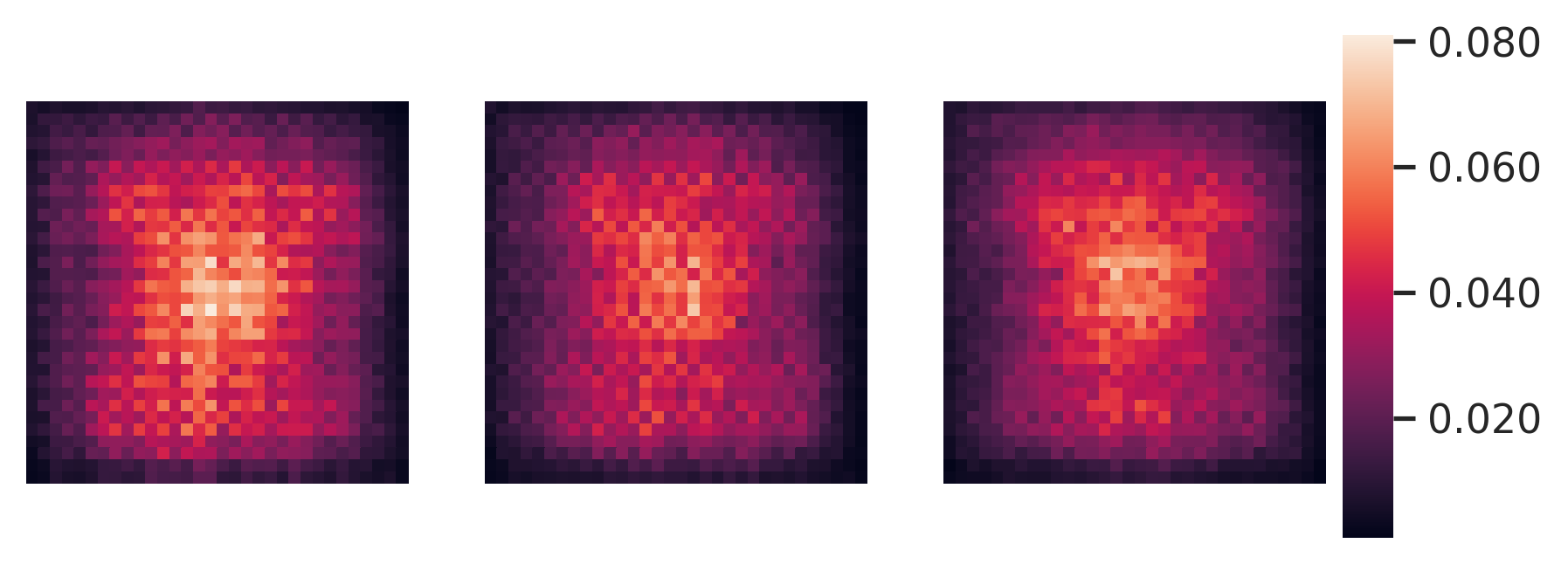}
    \end{subfigure}
    
    \caption{Heatmaps of $\hat{\tau}_i^\prime$ for the \textbf{Resnet101} architecture. Above: aggregated heatmap. Below: channel heatmaps.}
    \label{fig:heatmap_resnet}
    
\end{figure}

Figures \ref{fig:heatmap_lenet}, \ref{fig:heatmap_resnet} and \ref{fig:heatmap_densenet} show the heatmaps for the LeNet, Densenet and Resnet architectures, respectively. 
We first note that there are clear similarities for all networks: regions close to the border of the images are the least relevant, while the central regions tend to be important.  This is consistent with the nature of the training data, where the pixels important for the classification are typically found in the central region of the image. This general result is confirmed by the channel heatmaps, which show a similar structure for all color channels.

 One interesting difference for the different architectures is the extent of the border region which is assigned a near-zero importance: For LeNet, this region is considerably wider than for the more complex architecture DenseNet and Resnet, which could point to a more complex decision process within the latter models, taking more of the image into account.



\subsection{Mean Dimension Analysis}\label{sec:real_world_layer}
In this subsection, we report results for a layer-by-layer analysis of interaction studying a LeNet-5 architecture trained on the Cifar10 image classification data set. 
We restrict ourselves to LeNet-5 in this section since the very large depth of the DenseNet and ResNet architectures used by us would make a layer-by-layer analysis computationally very expensive.

Table \ref{cifar10_table} shows the LAMD of the trained network using the ReLU (mid column) and TanH (right column) activation functions. The left column shows the corresponding layer type to the LAMD. 
The final average mean dimension of the network with the TanH is larger than with the ReLU activation function. However, looking at the individual steps comparing the LAMD before and after activations, the increase of the LAMD is larger in all three cases for ReLU than TanH. See for example the increase in LAMD from row 1 to row 2, row 4 to 5 and row 7 to row 8.
\begin{table}[]
\begin{center}

\begin{tabular}{lll}
           & \multicolumn{2}{c}{\textbf{LAMD}} \\
\textit{Layer Type} &             &            \\
Conv2d     & 0.9991          & 0.9992          \\
ReLU/TanH  & 1.2382          & 1.1842          \\
MaxPool2d  & 1.0977          & 1.2396          \\
Conv2d     & 0.9019          & 1.1179          \\
ReLU/TanH  & 1.7292          & 1.7165          \\
MaxPool2d  & 3.2240          & 2.0922          \\
Linear     & 0.7864          & 1.1820          \\
ReLU/TanH  & 2.3429          & 1.3971          \\
Linear     & 0.8875          & 1.1484          \\
NLL        & 2.6912          & 3.1786         
\end{tabular}
\caption{\label{cifar10_table} LAMD of LeNet-5 type architecture trained on Cifar10 dataset using ReLU activation function (mid column) and hyperbolic tangent activation function (right column). The left column shows the corresponding layer type to the LAMD, where Conv2d is a 2D convolution layer, MaxPool2D is a 2D max pooling layer, linear is a fully-connected layer and Relu/TanH is the layer where we apply the activation function. NLL is the negative loglikelihood loss.}
    
\end{center}
\end{table}

\subsection{When Do Interactions Arise During Training?} \label{sec:real_world_training}
We now focus on evaluating the difference of artificial neural network architectures and on the evolution of the mean dimension during training on Cifar10.
We refer to the appendix for training details.
The estimated mean dimension for LeNet-5, ResNet-101 and DenseNet-121 are $2.97$, $7.91$ and $5.55$.  The training errors for all three models are zero and test errors for LeNet-5, ResNet-101 and DenseNet-121 are $29.55$, $19.18$ and $21.58$, respectively.
\begin{figure}
    \begin{subfigure}[b]{0.5\textwidth}
    \includegraphics[width=\textwidth]{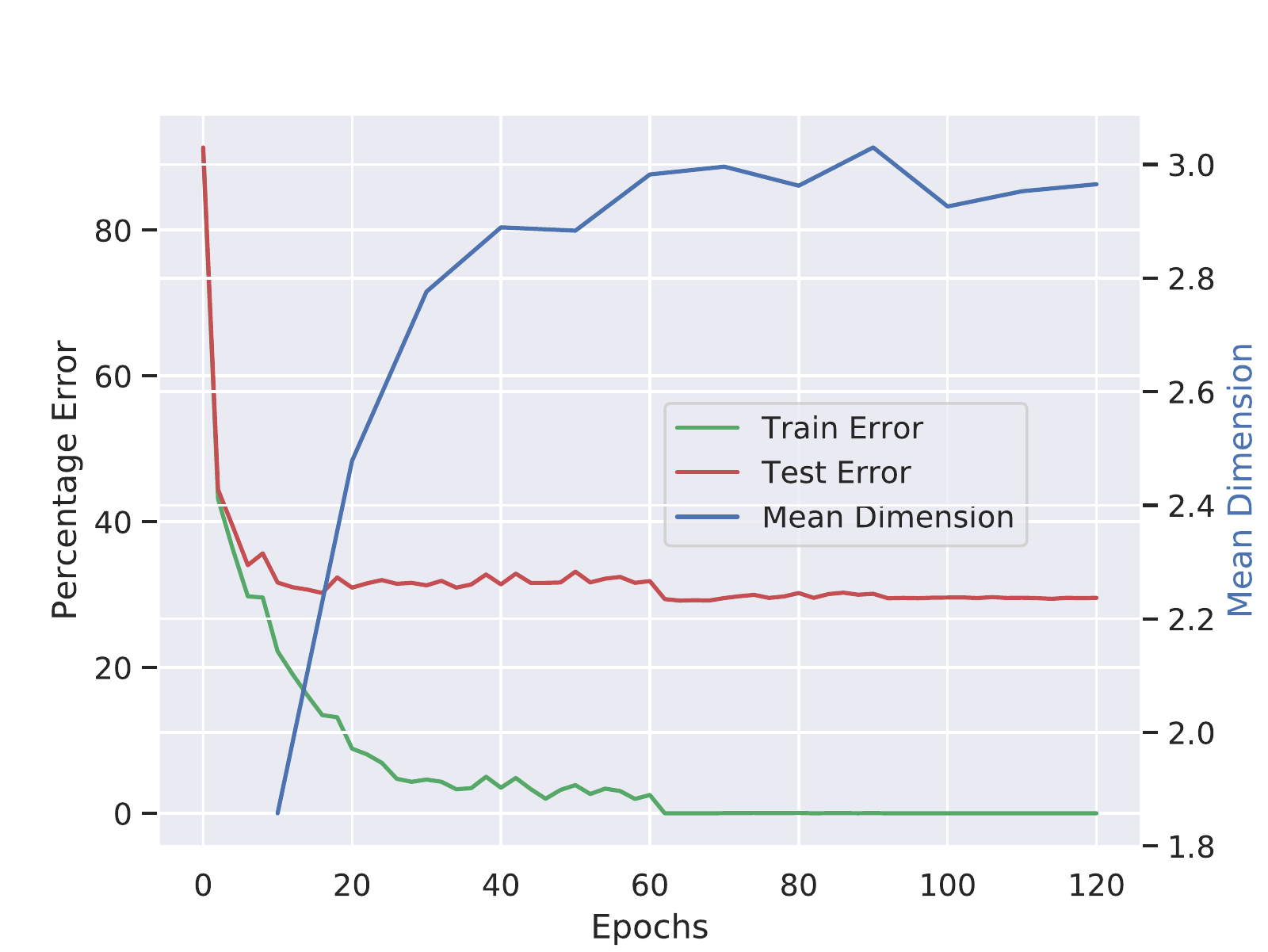}
    \caption{LeNet-5}
    \end{subfigure}
    \begin{subfigure}[b]{0.5\textwidth}
    \includegraphics[width=\textwidth]{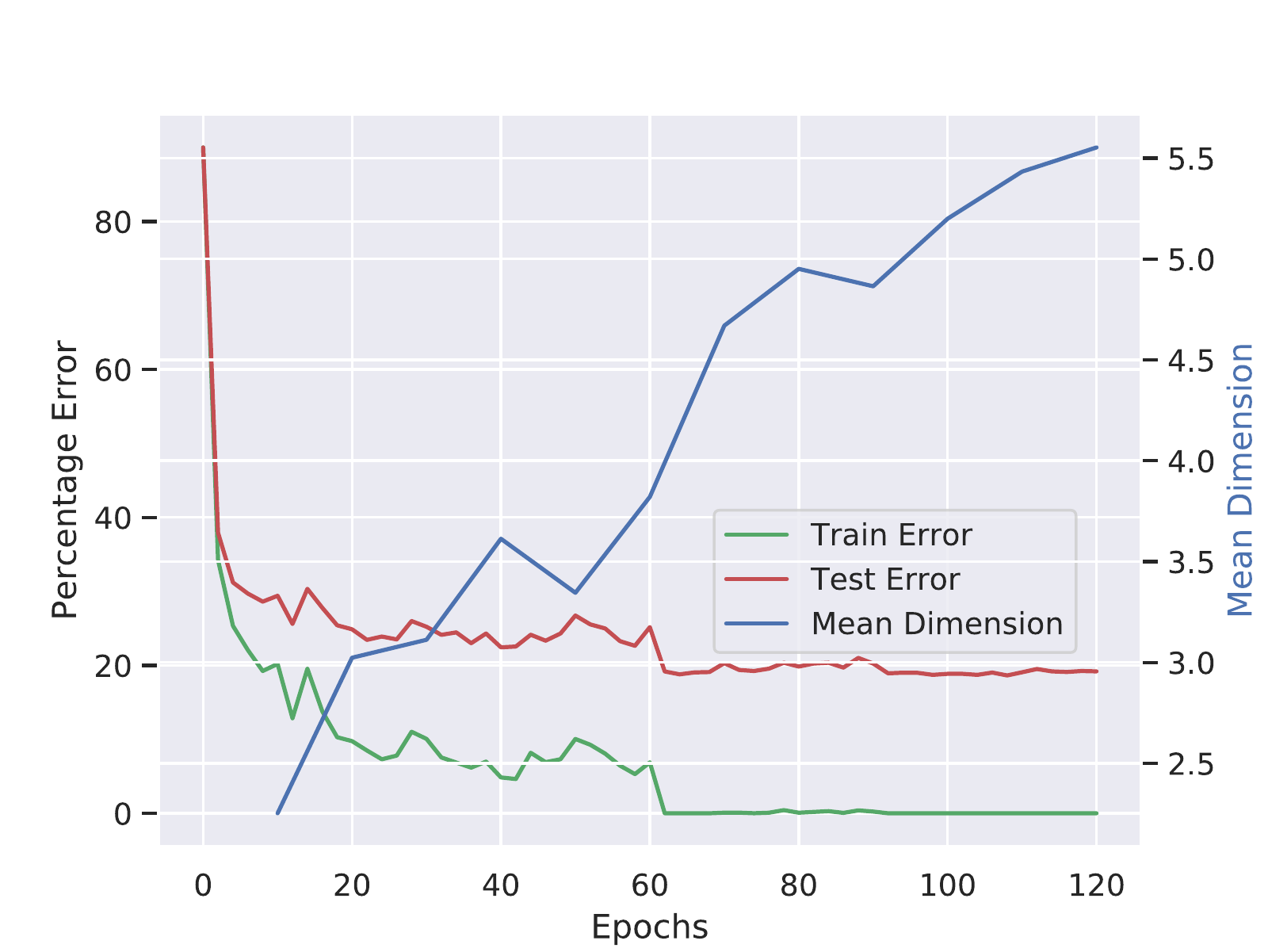}
    \caption{DenseNet-121}
    \end{subfigure} 

    \centering
    \begin{subfigure}[b]{0.5\textwidth}
    \includegraphics[width=\textwidth]{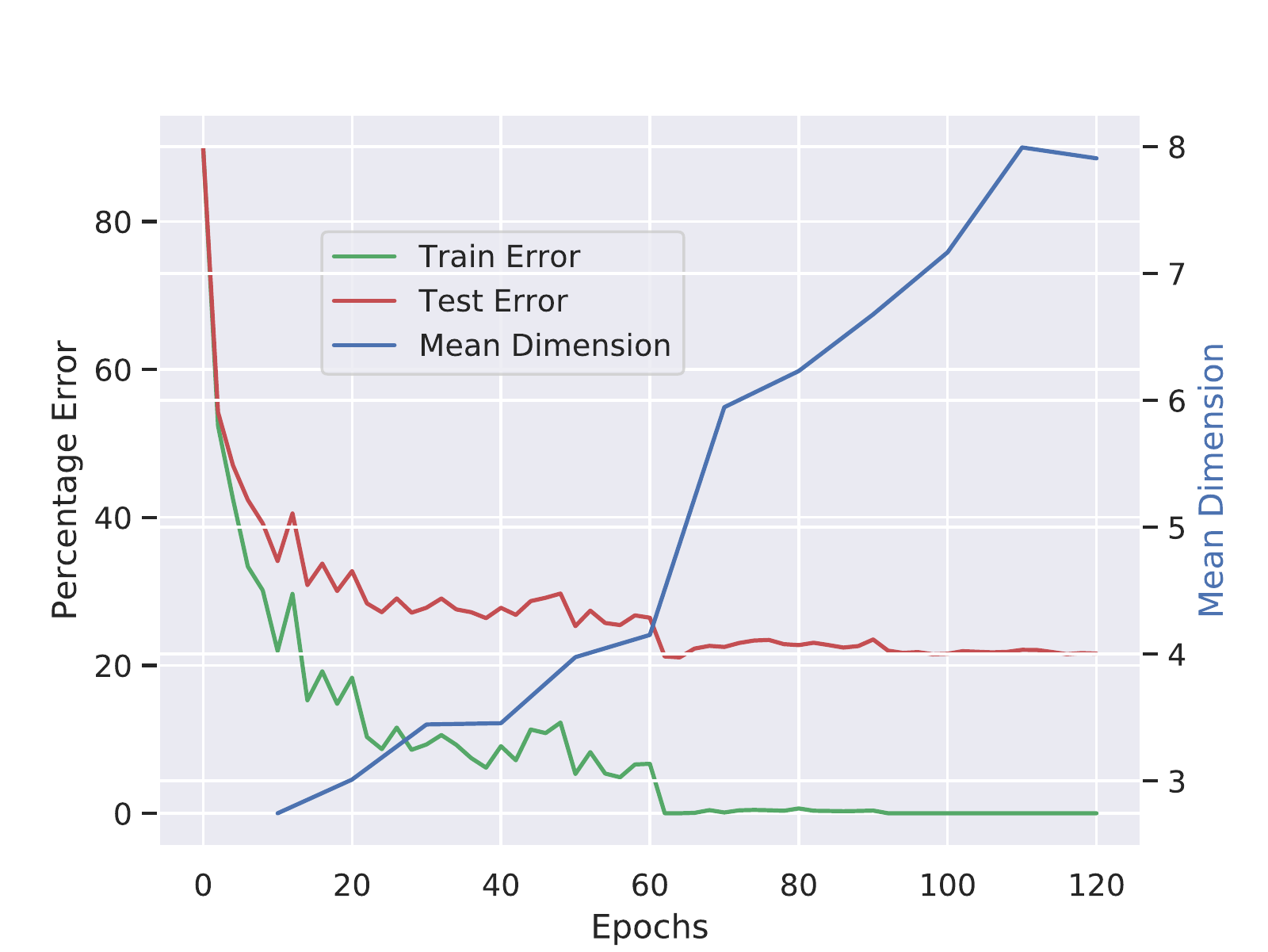}
    \caption{ResNet-101}
    \end{subfigure}    
    \caption{Mean dimension (MD, blue), training (green) and test error (red) evolution during training of 120 epochs using different architectures. The mean dimension is estimated every $10^{th}$ epoch of training. Errors are the percentage of misclassified images in the training and test set respectively.}
    \label{fig:diff_architectures}
\end{figure}

Figure \ref{fig:diff_architectures} displays the mean dimension estimates during training epoch by epoch from $10$ to $120$ epochs, as well as the train and test errors.  We observe that the mean dimension increases as the epochs increase.
 The increase in mean dimension is steeper for early epochs for the LeNet-5 architecture. The other two architectures do not seem to exhibit this pattern.
The deeper networks ResNet-101 and DenseNet-121 have a higher mean dimension than the more shallow LeNet-5.
This difference between architectures is visible from the beginning of training. We recall that there is one order of magnitude of difference in parameter size between the networks used in our experiment, with ResNet-101 having the largest number of layers followed by DenseNet-121 and then by LeNet-5. 
 On average, the mean dimension is more than 2.5 times higher in ResNet-101 than in LeNet-5 according to our estimates. Thus, the mean dimension seems to be positively correlated with the complexity of the network, with larger networks having larger mean dimension.   

The graphs in Figure \ref{fig:diff_architectures} may seem to indicate that the mean dimension converges more rapidly in the LeNet-5 model than in the other models. However, this is only due to the truncation at 120 epochs. We performed additional experiments with a higher number of epochs and the final mean dimension does not change substantially as the number of epochs increases also for ResNet-101 and DenseNet-121. That is, the least complex architecture LeNet-5 learns most of the interactions during the first epochs of training, while the other two architectures learn these interactions later on.

It is often true for computer vision tasks that a better accuracy can be achieved with bigger networks and more training, but for some problems these highly complex models might not be useful, for example due to overfitting. The mean dimension might give indications about when this is the case by showing that from a certain complexity on-wards the average interaction size stops increasing. Deciding how general this assertion is requires further experiments. These will be part of future research by the authors.

\begin{figure}
    \begin{subfigure}[b]{0.5\textwidth}
    \includegraphics[width=\textwidth]{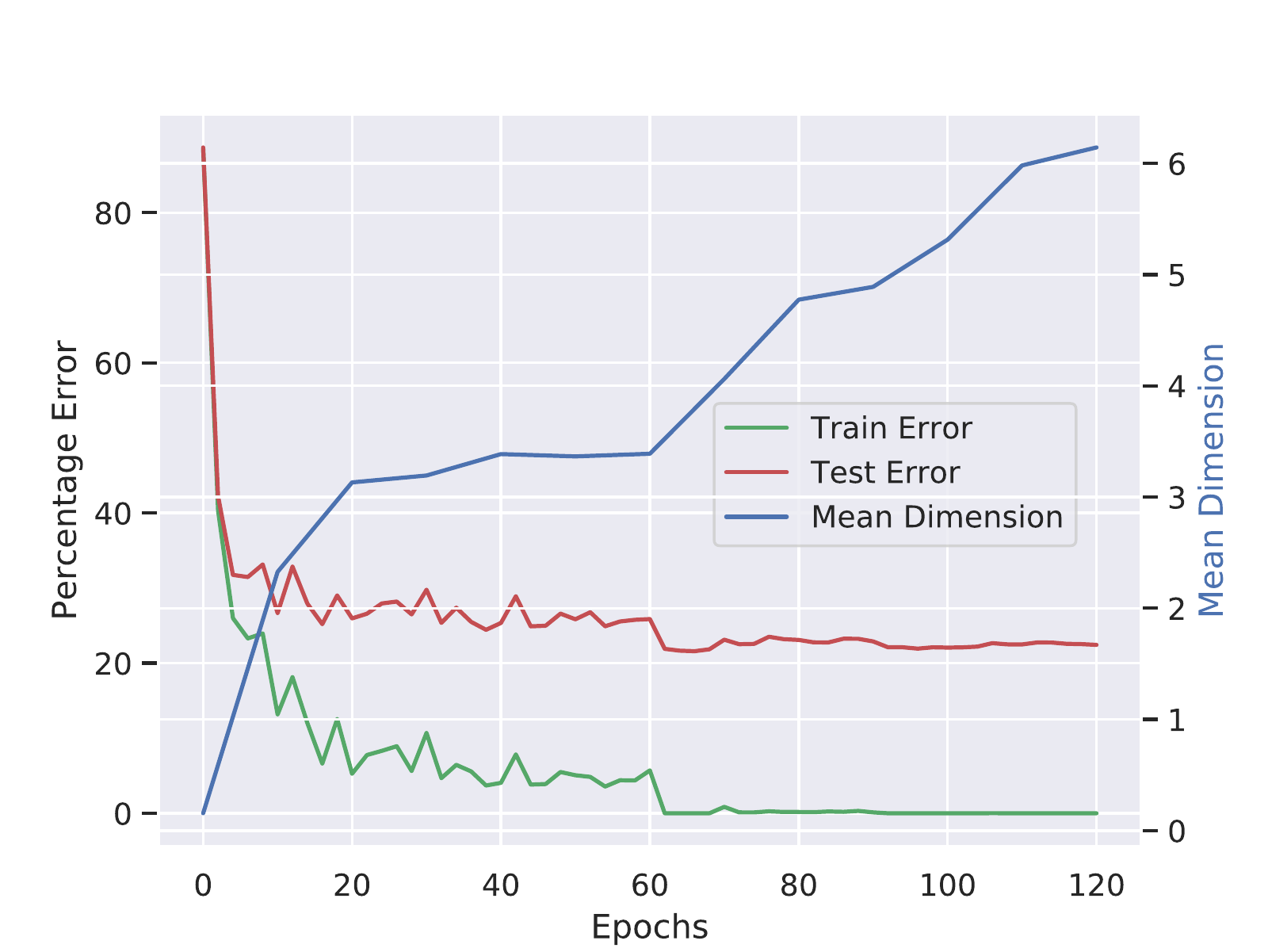}
    \caption{ResNet-18}
    \end{subfigure}
    \begin{subfigure}[b]{0.5\textwidth}
    \includegraphics[width=\textwidth]{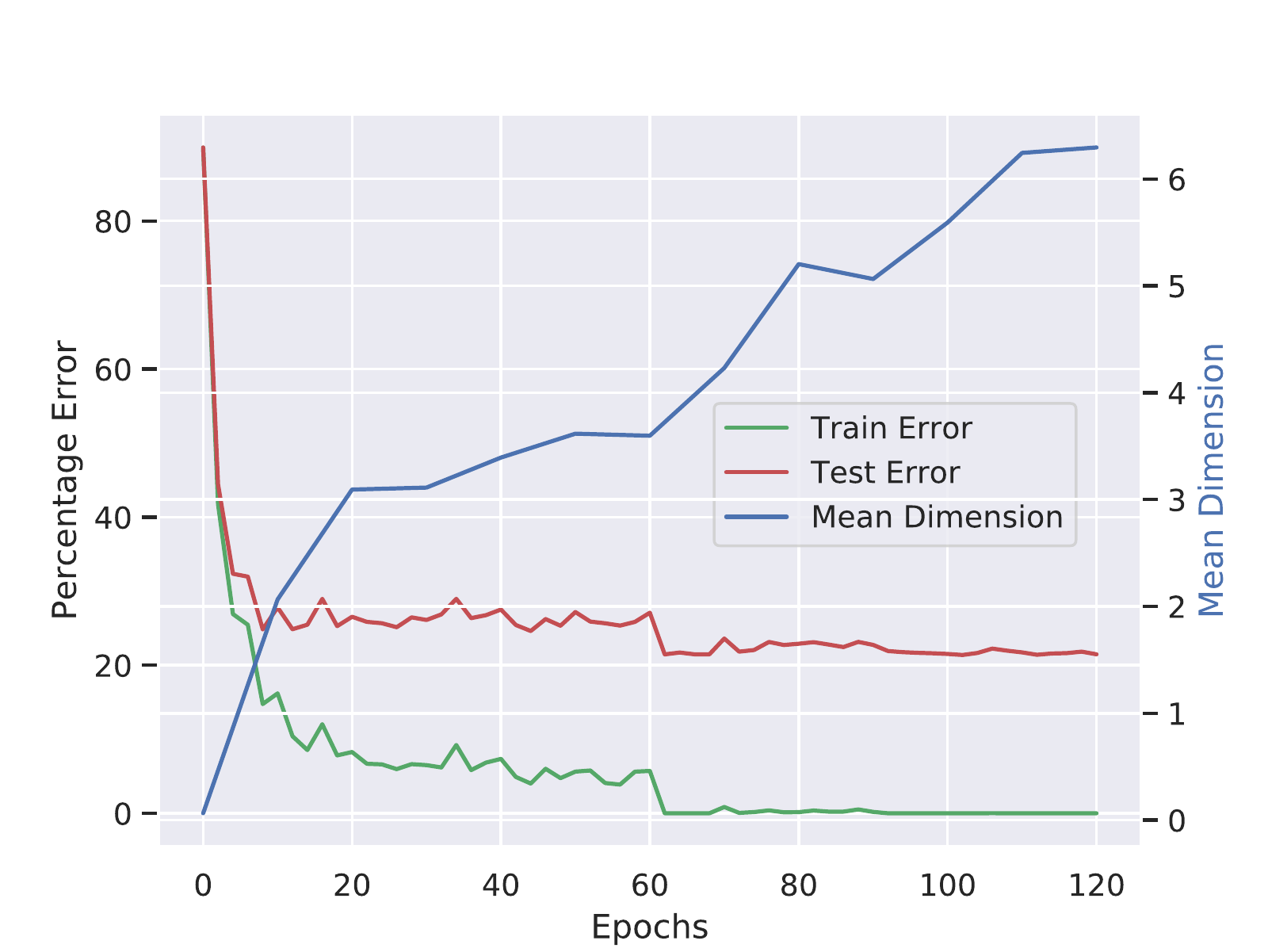}
    \caption{ResNet-34}
    \end{subfigure}
    
    \begin{subfigure}[b]{0.5\textwidth}
    \includegraphics[width=\textwidth]{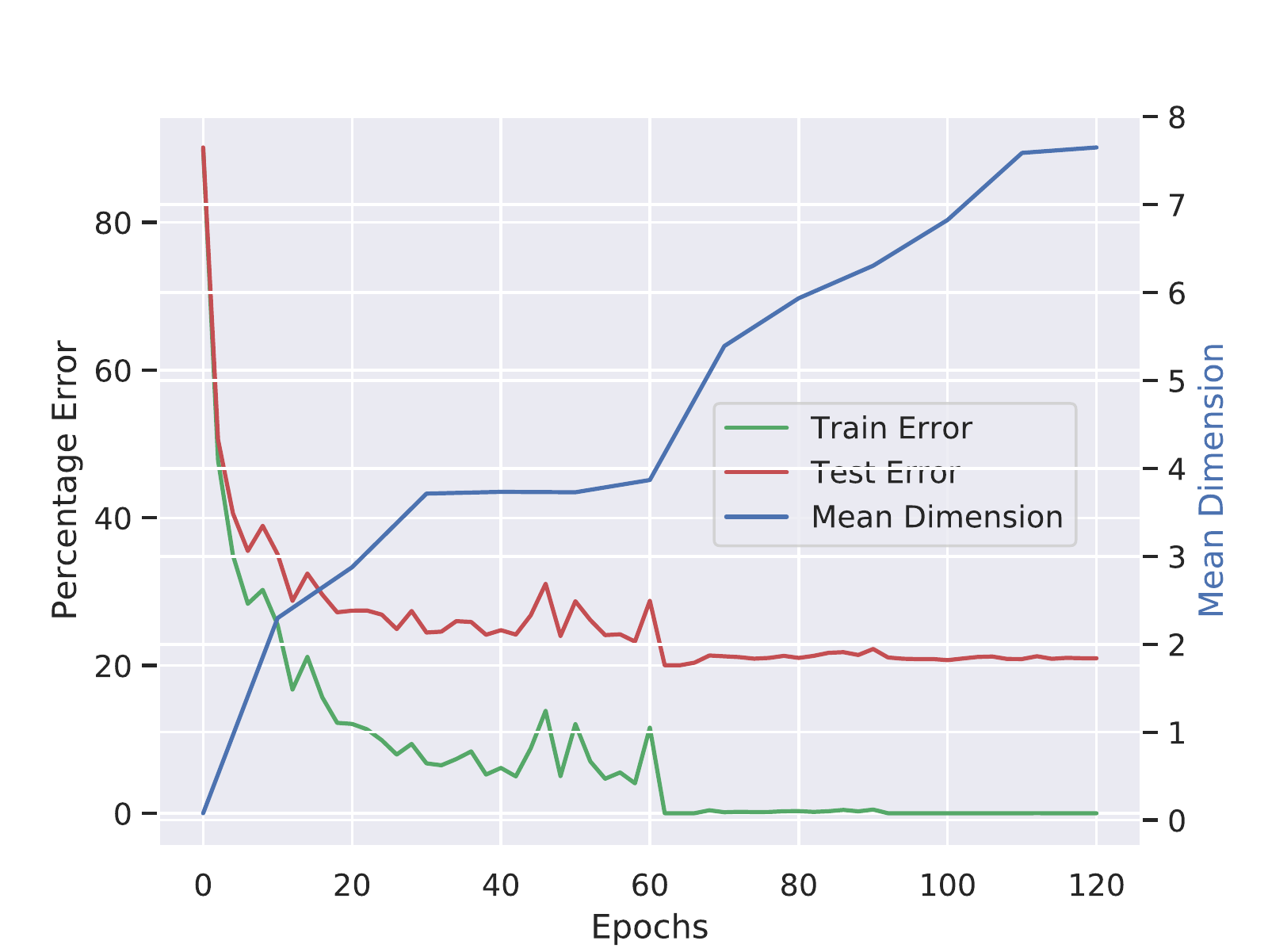}
    \caption{ResNet-50}
    \end{subfigure}
    \begin{subfigure}[b]{0.5\textwidth}
    \includegraphics[width=\textwidth]{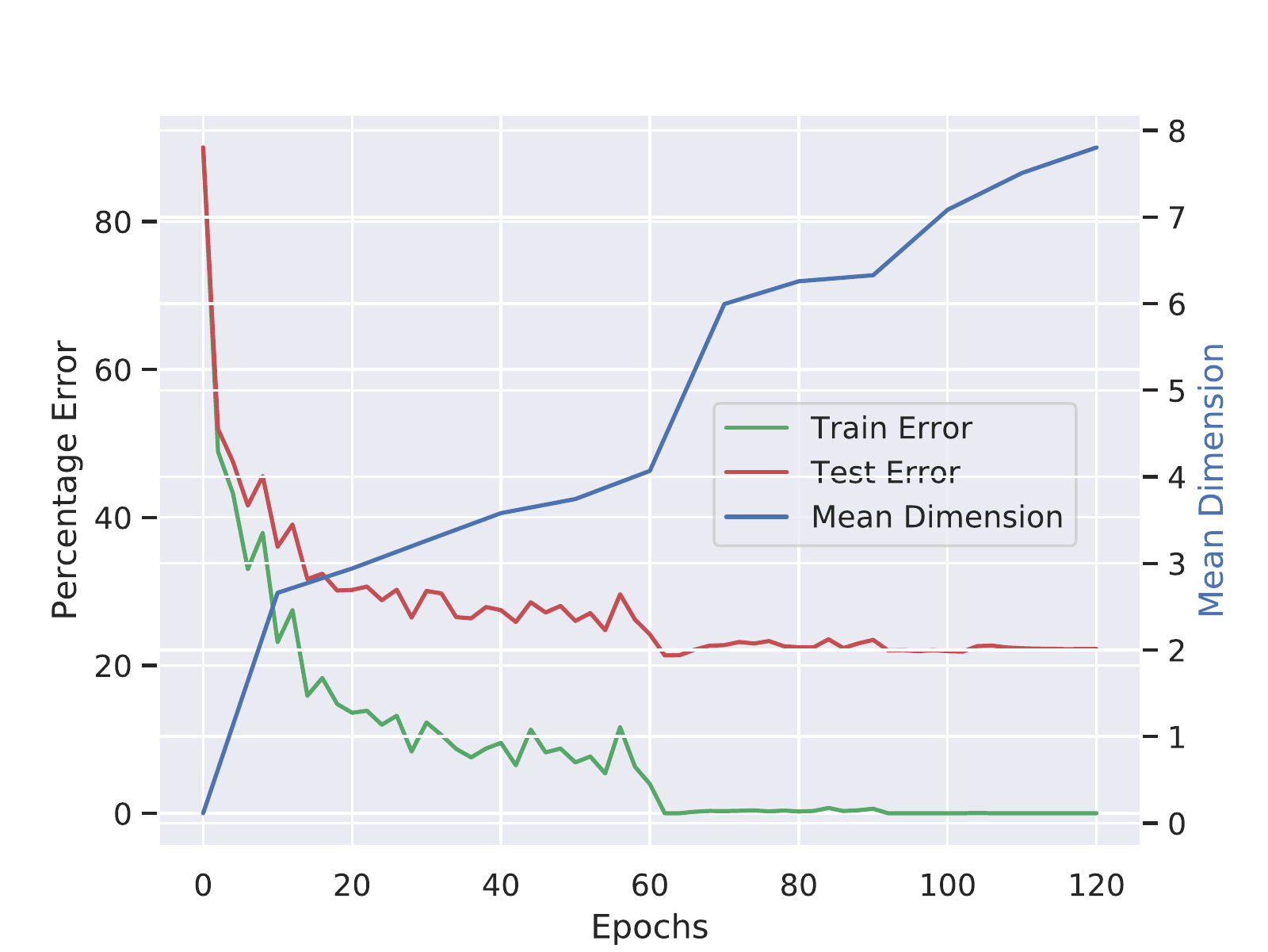}
    \caption{ResNet-152}
    \end{subfigure}
    \caption{Mean dimension (blue), training (green) and test error (red) evolution during training of 120 epochs using the ResNet architecture with 18 (a), 34 (b), 50 (c) and 152 (d) number of layers. The results using 101 layers (ResNet-101) is in Fig. \ref{fig:diff_architectures}.  The mean dimension is estimated every $10^{th}$ epoch of training.}
    \label{fig:resnet}
\end{figure}

In the next series of experiments, we estimate the mean dimension for the image classification task of Cifar10 using different versions of the ResNet architecture.
In Fig. \ref{fig:resnet}, we show the evolution of the mean dimension as well as training and test error for ResNet-18, ResNet-34, ResNet-50, and ResNet-152 as the number of epochs increases. The corresponding plot for ResNet-101 is in Fig. \ref{fig:diff_architectures}.
In accordance with the results of the previous experiments, we obtain larger mean dimension estimates for models with a higher number of layers. After 120 epochs these are $6.14$, $6.29$, $7.65$, $8.06$  and $7.80$, respectively, for ResNet-18, ResNet-34, ResNet-50, ResNet-101 and ResNet-152.

The analysis of the different depth versions of the ResNet architecture confirms that with increasing number of layers and neurons, an artificial neural network has a higher estimated mean dimension.
However, it is interesting to observe that after 50 layers, the mean dimension does not vary significantly. The mean dimension of ResNet-101 is only slightly higher and the mean dimension of ResNet-152 is even lower than the one of ResNet-101.
A similar behavior is observed in the test error, which is 22.43, 21.47, 20.98, 21.19, 22.20 for ResNet-18, ResNet-34, ResNet-50, ResNet-101 and ResNet-152, respectively.
One might conclude that ever increasing complexity does not lead to ever increasing mean dimension. There seems to be a bound for the mean dimension using the same architecture. Again this might indicate which complexity could be needed for a certain problem at hand.
Whether overfitting and an increase in mean dimension without an improvement in performance are related effects is an interesting direction for future research.

\section{Conclusion}\label{sec:discussion}
This work continues along the lines of the research on the mean dimension of artificial neural networks opened by \cite{OwenHoyt21}.
We have addressed the problem of estimating the mean dimension from a given dataset without resampling. We have then recorded the evolution of the mean dimension during training. Experiments have shown that the mean dimension increases with the number of training epochs, until it reaches a plateau of stable values. We have analyzed the behavior of the mean dimension through the various neurons and layers of the net, and performed experiments to appreciate the impact of the activation function. In particular, experiments have shown that ReLu activation functions cause the mean dimension to incur more sudden changes from layer to layer than the TanH activation function.

In order to compare deep network's architectures used in image classification, we have proposed the addition of an inverse PCA layer that preserves the input-output mapping structure but allows one to deal with uncorrelated features. 

The experiments have shown consistently that $D_g$ is lower than the overall problem dimensionality, a finding consistent with \cite{OwenHoyt21}. However, $D_g$ increases with the complexity of the architectures and, within the same architecture, with the number of layers. Interestingly, from a certain number of layers on-wards, the mean dimension stagnates.   

Finally, we have seen that with correlated features the design yields a variant of the total indices denoted with $\tau^\prime_i$, which are null if and only if the target is independent of the feature at hand. We have then examined the insights that these indices yield if employed as post-hoc explanations to produce heatmaps of pixel importance. 

This work also opens to future research directions. There are of course current limitations connected to, for example, the computational power available for our experiments. Repeating the calculations for larger ensemble of networks and analyzing the mean dimension in relation to generalization, similarly to the idea of \cite{jiang2019fantastic} is a further avenue of future research.


\FloatBarrier

\bibliographystyle{siamplain}
\bibliography{references,library}

\end{document}